\documentclass[preprint,authoryear,12pt]{elsarticle}

\usepackage{times}
\usepackage{helvet}
\usepackage{courier}

\usepackage{algorithmic}
\usepackage{amsmath}
\usepackage{amssymb}
\usepackage{amsthm}
\usepackage{latexsym}
\usepackage{epsfig}
\usepackage{alltt}
\usepackage{url}
\usepackage{graphicx}
\usepackage{float}
\usepackage{multirow}
\usepackage{color}
\usepackage{fullpage}
\usepackage[ruled,linesnumbered,noresetcount]{algorithm2e}
\usepackage[english]{babel}
\usepackage[autostyle]{csquotes}
\usepackage[kerning,spacing,protrusion,babel,final]{microtype}

\usepackage{tikz}
\usetikzlibrary{arrows,chains,positioning,automata,decorations,shapes,calc,matrix,fit} 

\allowdisplaybreaks

\long\def\COMMENT#1\ENDCOMMENT{\message{(Commented text...)}\par}

\newboolean{includeMemo}
\setboolean{includeMemo}{true}

\newcommand{\memo}[1]{%
    \ifthenelse{\boolean{includeMemo}}{%
        \medskip\noindent\fbox{%
            \begin{minipage}[b]{%
                \dimexpr\linewidth-1em}#1\end{minipage}
            }
        \medskip\newline
    }
}

\usepackage{aij-notation}
\usepackage[colorinlistoftodos,prependcaption,textsize=small]{todonotes}

\newtheorem{definition}{Definition}
\newtheorem{theorem}{Theorem}

\newtheorem{example}{Example}
\newtheorem{remark}{Remark}
\def\qed{\relax\ifmmode\hskip2em \Box\else\unskip\nobreak\hfill $\Box$\smallskip\fi}

\def\calai{{\cal AI}}
\allowdisplaybreaks

\journal{AI Journal}

\begin{document}



\begin{frontmatter}
\title{
    An Action Language for Multi-Agent Domains
}

\author{
    Chitta Baral$^\dagger$,
    Gregory Gelfond$^\ddagger$,
    Enrico Pontelli$^\diamond$,
    Tran Cao Son$^{\diamond \star}$
}

\ead{
    chitta@asu.edu,
    ggelfond@unomaha.edu,
    epontell@cs.nmsu.edu,
    tson@cs.nmsu.edu
}

\address{
    $^\dagger$Department of Computer Science and Engineering, Arizona State University\\
    $^\ddagger$Department of Computer Science, University of Nebraska Omaha\\
    $^\diamond$Department of Computer Science, New Mexico State University\\
    $^{\star}$Corresponding author \\
}

\begin{abstract} 

The goal of this paper is to investigate an action language, called \mastar{}, for representing and reasoning about actions and change in multi-agent domains. 
The language, as designed, can also serve as a specification language for epistemic planning, thereby addressing an important issue in the development of multi-agent epistemic planning systems. 
The \mastar{} action language is  a generalization of the single-agent action languages, extensively studied in the literature,   to the case of \emph{multi-agent domains.} The language allows the representation of different types of  actions that an agent can perform in a domain where many other agents might be present---such as world-altering actions, sensing actions, and communication actions.  The action language also allows  the specification of agents'  \emph{dynamic awareness} of  action occurrences---which has  implications on what agents' know about the world and other agents' knowledge about the world. These features are embedded in a language that is simple,  yet powerful enough to address a large variety of knowledge manipulation scenarios in multi-agent domains.

The semantics of \mastar{} relies on the notion of \emph{state,} which is described by a  \emph{pointed Kripke model} and is used to encode the agents' knowledge\footnote{We will use the term ``knowledge'' to mean both ``knowledge'' and ``beliefs'' when  clear from the context.} and the real state of the world. The semantics is defined by a transition function that maps pairs of actions and states  into sets of states. The paper presents a number of properties of the action theories and relates \mastar{} to other relevant formalisms in the area of reasoning about actions in multi-agent domains.

\end{abstract}


\end{frontmatter}




\section{Introduction}

\subsection{Motivations}

\emph{Reasoning about Actions and Change (RAC)} has been a research focus since the early days of artificial
intelligence \citep{mccarthy59}. Languages for representing actions and their effects 
have been proposed soon after \citep{fik71}. While the early papers on this topic by \cite{fik71} did not include formal semantics, papers  with formal semantics came some years after, with leading efforts by \cite{lif87c}. 
The approach adopted in this paper is predominantly  influenced by
the methodology for representing and reasoning about actions and change
proposed by 
 \cite{GelfondL93}. In this approach, actions of agents are described in a high-level language, with an English-like syntax and a transition 
 function-based semantics. Action languages offer several benefits, including  a succinct way for representing dynamic domains. The approach 
 proposed in this paper is also related to action description languages developed for planning, such as \citep{ped89,GhallabHKM98}.
 
Over the years,  several action languages (e.g., $\cal A$, $\cal B$, and $\cal C$) have been  developed,
 as discussed in \citep{GelfondL98}. Each of these languages addresses some important problems in RAC,  e.g., the ramification problem, concurrency, actions with duration, 
knowledge of agents. Action languages have also provided the foundations for several successful approaches to automated 
planning; for example, the language $\cal C$ is used in the planner {\sc C-Plan} \citep{CastelliniGT01} and the language 
$\cal B$ is used in  {\sc CpA} \citep{SonTGM05a}. 
The \emph{Planning Domain Definition Language (PDDL)}  \citep{GhallabHKM98}, the de-facto 
standard language for planning systems, could also be viewed as a type of action language (a generalization of STRIPS).
However, the primary  focus of all these research efforts  has been about reasoning within
 \emph{single-agent domains.} 

In single-agent domains, reasoning about actions and change mainly involves reasoning about what is true in the world, what the agent knows about the world, how the agent can manipulate the world (using world-changing actions) to reach particular states, and how the agent (using sensing actions) can learn unknown aspects of the world. In  \emph{multi-agent domains} an agent's action may not just change the world and the agent's knowledge about the world, but also may change  other agents' knowledge. Similarly, 
the goals of an agent in a multi-agent world may involve manipulating the knowledge of other agents---in particular, this may involve
 not just their knowledge about the world, but also their knowledge about other agents' knowledge about the world. Although there is a large body of research on multi-agent planning (see, e.g., \citep{Durfee99,WeerdtBTW03,WeerdtC09,AllenZ09,BernsteinGIZ02,GoldmanZ04,GuestrinKP01,NairTYPM03,PeshkinS02}), 
 relatively  few efforts address the above aspects of multi-agent domains,  which offer a number of new research challenges in representing and reasoning about actions and change. The following simple example illustrates some of these issues. 

\begin{example}
[Three Agents and the Coin Box]
 \label{ex1} 
{  
Three agents, $A$, $B$, and $C$, are in a room. In the middle 
of the room there is a box containing a coin. It is common knowledge 
that: 
\begin{list}{$\bullet$}{\topsep=1pt \parsep=0pt \itemsep=1pt}
\item None of the agents knows whether the coin lies heads or tails up;

\item The box is locked and one needs  a key to open it; 
	 agent $A$ has the key of the box and everyone knows this;  

\item In order to learn  whether the coin lies heads or tails up, an
agent can peek into the box---but this requires the box to be
open; 

\item If one agent is looking at the box and a second agent peeks into the box, 
then the first agent will observe this fact and will be able to conclude that 
the second agent knows the status of the coin; on the other hand, 
the first agent's knowledge about which face of the coin is up 
does not change;

\item Distracting an agent causes that agent  to not look at the box; 

\item Signaling an agent to look at the box causes such agent to 
look at the box; 

\item Announcing that the coin lies heads or tails up will make this a common knowledge
among the agents that are listening.
\end{list}
Suppose that the agent $A$ would like to know whether the coin lies 
heads or tails up. She would also like to let the agent $B$ know
that she knows this fact. However, she would like to keep this 
information secret from $C$. Please, observe that 
 the last two sentences express goals that are about agents' knowledge about other agents' knowledge. 
 Intuitively, she could achieve
her goals by:
\begin{enumerate}
\item Distracting $C$ from looking at the box; 
\item Signaling $B$ to look at the box if $B$ is not looking at the box; 
\item Opening the box; and 
\item Peeking into the box.\hfill$\Box$ 
\end{enumerate}
}
\end{example} 
This simple story presents a number of challenges for research in 
representing and reasoning about actions and their effects in 
multi-agent domains. In particular:

\begin{itemize} 
\item 
The domain contains several types of actions: 

\begin{itemize} 

\item Actions that allow the agents to change the state of 
the world (e.g., opening the box);

\item Actions that change the knowledge of the agents 
(e.g, peeking into the box, announcing heads/tails);

\item Actions that manipulate the beliefs of other agents 
(e.g., peeking while other agents are looking); and 

\item Actions that change the observability of agents with respect to awareness about future 
actions (e.g., distract and signal actions before peeking into the box). 

\end{itemize}
We observe that the third and fourth types of actions 
are not considered in single agent systems. 
\item The reasoning process that allows agent $A$ to verify that 
steps (1)-(4) will indeed achieve her goal requires $A$'s ability to  
reason about the effects of  actions  on several aspects:
\begin{itemize} 
\item \emph{The state of the world}---e.g., opening the box 
causes the box to become open;

\item \emph{The agents' awareness of the environment and of other
agents' actions}---e.g., distracting or signaling an agent 
causes this agent not to look or to look at the box, respectively;   
and

\item \emph{The knowledge of other agents about her own knowledge}---e.g., someone 
following her actions would know what she knows. 
\end{itemize}
While the first requirement is the same as for an agent in single-agent domains, 
the last two are specific to multi-agent domains. 
\end{itemize}
With respect to planning, the above specifics of multi-agent 
systems raise an interesting problem: 

{\it\begin{quote}
``How can
one generate a plan for the agent $A$ to achieve her goal, 
given the description in Example~\ref{ex1}?''
\end{quote}}

The above problem is an \emph{Epistemic Planning problem 
in a Multi-agent domain (EPM)} \citep{BolanderA11} which refers to the
generation of plans for multiple agents to achieve goals which can refer to the
state of the world, the beliefs of agents, and/or the knowledge of agents. 
EPM has recently attracted the attention of researchers from various communities
such as planning, dynamic epistemic logic, and knowledge representation. The
Dagstuhl seminars on the subject \citep{ALLB14,BBMvD17} provided the impetus for
the development of several epistemic planners
\citep{KominisG15,HFWL17,Muise+15,WanYFLX15,LiuL18,LeFSP18} and 
extensive studies of the
theoretical foundations (e.g., decidability and computational complexity)
of EPM \citep{AucherB13,BolanderJS15}. 
In spite of all these efforts, to the best of our knowledge, 
only two systems have been proposed that address the complete range of
issues mentioned in Example~\ref{ex1}:
the dynamic epistemic 
modeling system called DEMO \cite{Eijck04} and the recently proposed system 
described in  \cite{LeFSP18}.
This is in stark contrast to 
the landscape of automated planning for single-agent domains, 
where we can find several efficient automated planners capable of generating 
plans consisting of hundreds of actions within seconds---especially
building on recent advances in search-based planning.

Among the main reasons for the lack of planning systems 
capable of dealing with the issues like those shown in  Example~\ref{ex1} are: 
{\bf (i)} the lack of action-based formalisms that can address the above mentioned 
issues and that can actually be orchestrated, and 
{\bf (ii)}  the fact that logical approaches to reasoning about 
knowledge of agents in multi-agent domains are mostly 
model-theoretical, and  not  amenable to an implementation 
in search-based planning systems. Indeed, both issues were raised in the recent 
Dagstuhl seminar \citep{BBMvD17}. The issue
{\bf (i)} is considered as one of the main research topics in EPM,
while  
{\bf (ii)} is related to the practical and conceptual knowledge representation
challenges---discussed by Herzig\footnote{\url{http://materials.dagstuhl.de/files/17/17231/17231.AndreasHerzig.Slides.pdf}}
at the second Dagstuhl seminar \citep{BBMvD17}. We will discuss these issues in more detail 
in the next sub-section.  
 

\subsection{Related Work}
In terms of related work,  multi-agent actions have been explored  in 
\emph{Dynamic Epistemic Logics (DEL)} (e.g., \cite{BaltagM04,HerzigLM05,vBenthem07,BenthemEK06,vanDitmarschHK07}). However, as discussed later in the paper,  DEL does not offer an intuitive view of
how to  orchestrate or execute a single multi-agent action. 
In addition, the complex representation of multi-agent actions---similar to a Kripke structure---drastically increases the number of possible multi-agent actions---thus,  making it challenging to adopt a search-based approach in
developing   multi-agent action sequences to reach a given goal. It can be observed 
that several approaches to epistemic planning in multi-agent domains with focus on knowledge and beliefs of agents did 
employ an extension of PDDL rather than using DEL \citep{KominisG15,HFWL17,Muise+15,WanYFLX15,LiuL18}.  


The research in DEL  has also not addressed
some critical aspects of multi-agent search-based planning, such as the determination of the initial state of a planning domain instance.
Moreover, research in DEL did not explore the link between the state of the world 
and the observability encoded in multi-agent actions, and hence preventing  the dynamic evolution of  the observational capabilities and awareness of the agents with respect to future actions.  In some ways, the DEL approach is similar to the formulation of belief updates (e.g.,
\citep{update1,kat92,delval94}), and most of the differences and similarities between belief updates and reasoning about actions carry over to the differences and similarities between DEL and our formulation of RAC in multi-agent domains. We will elaborate on these differences in a later section of the paper.

\subsection{Contributions and Assumptions}

Our goal in this paper is to develop a framework that allows reasoning about actions and their effects in a multi-agent domain; the framework is expected to address  the above-mentioned issues, e.g., actions' capability
to modify  agents' knowledge and beliefs about  other agents' knowledge and beliefs. To this end, we propose a high-level action language for  representing and reasoning about actions in multi-agent domains. The language provides the
fundamental components of a planning domain description language for multi-agent systems.
The main contributions of the paper are:

\begin{itemize}

\item The action language \mastar{}, which 
allows the representation of different types of actions---such as world-altering actions,
announcement actions, and sensing actions---for formalizing multi-agent domains; the 
language explicitly supports actions that allow the dynamic
modification of the  awareness and observation
capabilities of the agents;

\item A transition function-based semantics for \mastar{}, that enables 
 hypothetical reasoning and planning in multi-agent domains. This, together with the notion of a 
 \emph{finitary-{\bf S5} theories} 
 for representing the initial state, introduced  in \citep{SonPBG14}, provides a 
 foundation for  the 
 implementation of a heuristic search-based planner for
  domains described in \mastar{}; and

\item Several theoretical results relating  the semantics of {\mastar{}} to multi-agent actions
characterizations using  the notion of update models from DEL  \cite{BaltagM04}.

\end{itemize}
In developing \mastar{}, we  make several  design decisions. The key decision is that actions in our formalism can  be effectively executed and the outcome can be effectively determined. This is not the case, for example,  in DEL \citep{vanDitmarschHK07}, where actions are complex graph structures, similar to Kripke structures,  possibly representing a multi-modal formula, and it is not clear if and how such actions can be executed. We also assume that actions are deterministic, i.e., the result of the execution of a world altering action is unique. This assumption can be lifted in a relatively simple manner---by generalizing 
the techniques for handling non-deterministic actions studied in the context of  single-agent domains. 

Although we have mentioned both knowledge and beliefs, in this paper we will follow  \cite{vanDitmarschHK07,BaltagM04} and focus only on formalizing the changes of \emph{beliefs} of agents after the execution of actions. Following the considerations by \cite{vBenthem07}, the 
epistemic operators used in this paper can be read as \emph{``to the best of my information.''}
Note that, in a multi-agent system, there may be a need to distinguish between 
the \emph{knowledge} and the \emph{beliefs} of an agent about the world. Let us 
consider Example~\ref{ex1} and let us denote with $p$ the proposition 
\emph{``nobody knows whether the coin lies heads or tails up.''} Initially, the three agents know that $p$ is true. However, after agent $A$ executes the  sequence of actions (1)-(4), $A$ will know that $p$ is false. Furthermore, $B$ also knows that $p$ is false, thanks  to her awareness of $A$'s execution of the actions  of opening the box and looking into it. However, $C$, being unaware of the execution of the actions
performed by $A$, will still believe that $p$ is true. 
If this were considered as a part of $C$'s knowledge, then $C$ would result in having false knowledge. 

The investigation of the
discrepancy between knowledge and beliefs has been an intense 
research topic in dynamic epistemic logic and in reasoning about knowledge, 
which has lead to the development of several 
modal logics (e.g., \citep{FaginHMV95,vanDitmarschHK07}).  Since our main 
aim is the development of an action language for hypothetical reasoning and planning, 
we will be primarily concerned  with the beliefs of agents. 
Some preliminary steps in this direction have been explored in the context
of the DEL framework  \citep{HerzigLM05,SonPBG15}. We leave the development of an action-based formalism  
that takes into consideration the differences between beliefs and knowledge as future work.

%


\subsection{Paper Organization}
The rest of the paper is organized as follows. Section \ref{background} reviews the 
basics definitions and notation of a modal logic with belief operators and the update model based approach 
to reasoning about actions in multi-agent domains. 
This section also reviews the definition of finitary {\bf S5}-theories whose models are finite.
It also includes a short discussion for the development of \mastar{}.
Section \ref{syntax} presents the syntax of \mastar{}. Section \ref{semantics} 
explores the modeling of the semantics
of \mastar{} using the update models approach; we define the transition function 
of \mastar{} which maps pairs of actions and states into states; the section also
presents 
the entailment relation between \mastar{} action theories
and queries along with relevant properties.
Section \ref{related} provides an analysis of \mastar{} with respect to the existing literature, including a comparison with  DEL.
 Section \ref{conclusion} provide some concluding remarks and directions for 
  future work. 
   For simplicity of  presentation, the proofs of the main theorems are placed in Appendix A.


\section{Preliminaries} 
\label{background} 


We begin with a review of the basic notions from the literature on 
formalizing knowledge and reasoning about effects of actions in 
multi-agent systems. Subsection~\ref{sub-belief} presents 
the notion of Kripke structures. Subsection~\ref{sub-update} 
reviews the notion of update models developed by the dynamic 
epistemic logic community for reasoning about effects of actions 
in multi-agent systems. 


\subsection{Belief Formulae and Kripke Structures}
\label{sub-belief}

Let us consider an environment with a set $\agents$ of $n$ agents.
The \emph{real state of the world} (or \emph{real state}, for brevity) 
is described by a set $\calf$ of 
propositional variables, called {\em fluents}. We   
are concerned with the beliefs of agents about 
the world and about the beliefs of other agents. 
For this purpose, we adapt the 
logic of knowledge and the notations used in~\cite{FaginHMV95,vanDitmarschHK07}. 
We associate to each agent $i \in \agents$ a modal operator $\B_i$ (to indicate
a belief of agent $i$) and  
represent the beliefs of an agent as belief formulae in a logic extended 
with these operators. Formally, we define belief formulae using the BNF: 
\[
  \varphi {:}{:} =   p \mid \neg \varphi \mid (\varphi \wedge \varphi) \mid (\varphi \vee \varphi)  \mid \B_i\varphi \mid \mathbf{E}_\alpha \varphi \mid \mathbf{C}_\alpha \varphi
\]
where $p$ is a fluent, $i \in \agents$, and $\emptyset \neq \alpha \subseteq \agents$. We often use \emph{a fluent formula} 
to refer to a belief formula which does not contain any occurrence of $\B_i$, $\mathbf{E}_\alpha$, and $\mathbf{C}_\alpha$.

%
%
%
%

Formulae of the form $\mathbf{E}_\alpha \varphi$ and $\mathbf{C}_\alpha \varphi$ 
are referred to as \emph{group formulae.} Whenever $\alpha = \agents$, we simply write 
$\mathbf{E} \varphi$ and $\mathbf{C} \varphi$ to denote
$\mathbf{E}_\alpha \varphi$ and $\mathbf{C}_\alpha \varphi$, respectively. 
Let us denote
with $\call_\agents$ the language of the belief formulae over $\calf$ and \agents.

Intuitively, belief formulae are used to describe the beliefs of one agent 
concerning the state of the world as well as about the beliefs of other agents. 
For example, the formula $\B_1 \B_2 p$ expresses the fact that 
    ``{\em Agent 1 believes that agent 2 believes that $p$ is true},'' while 
$\B_1 f$ states that ``{\em Agent 1 believes that $f$ is true}.''   

In what follows, we will simply talk about  ``formulae'' instead of 
``belief formulae,'' whenever there is no risk of confusion. In order 
to define the semantics of such logic language, we need to introduce 
the notion of a Kripke structure. 
\begin{definition}
[Kripke Structure]
\label{definition-kripke-structure}
{
A \textit{Kripke structure} is a tuple $\langle S,\pi, \RK_{1},\ldots,\RK_{n} \rangle$, 
where 
\begin{itemize} 
\item $S$ is a set of worlds, 
\item $\pi: S \mapsto 2^\calf$ is a function that associates an interpretation of $\calf$ to each 
element of $S$, and 
\item  For $1 \leq i \leq n$, $\RK_{i} \subseteq S \times S$ is a binary relation over $S$.
\end{itemize} 
A \emph{pointed Kripke structure}  
is  a pair $(M,s)$ where 
$M = \langle S,\pi,\RK_{1},\ldots,\RK_{n} \rangle$ is
a Kripke structure and $s \in S$. In a pointed Kripke structure $(M,s)$, we
refer to $s$ as the \emph{real} (or \emph{actual}) world.
}
\end{definition}
   For the sake of readability, we use $M[S]$, $M[\pi]$, and $M[i]$ to denote the components 
$S$, $\pi$, and $\RK_{i}$ of $M$, respectively. For $u \in S$,  we write $M[\pi](u)$ to denote the interpretation associated to $u$ via $\pi$ 
and  $M[\pi](u)(\varphi)$ to denote the truth value of a fluent formula 
$\varphi$ with respect to the interpretation $M[\pi](u)$. 
In keeping with the tradition of action languages, we will 
  often refer to $M[\pi](u)$ as the set of fluent literals 
  defined by\footnote{
     For simplicity of the presentation, we often omit the negative literals as well.  
  } 
  $$
  \{f \mid f \in \fluents, M[\pi](u)(f) = \top\} \cup \{\neg f \mid f \in \fluents, M[\pi](u)(f) = \bot\}.
  $$ 
  Given a consistent and complete set of literals $X$, i.e., 
  $|\{f, \neg f\} \cap X| = 1$ for every $f \in \fluents$, 
  we write   $M[\pi](u) = X$ to indicate   
  that the interpretation $M[\pi](u)$ is defined in such a way that $M[\pi](u) = X$.

   Intuitively, a Kripke structure describes the possible worlds envisioned by the agents---and the presence
  of multiple worlds identifies uncertainty and the existence of different beliefs. The relation 
  $(s_1, s_2)\in \RK_i$ denotes that the belief of agent $i$ about the
  real state of the world is insufficient for her to distinguish between 
the world described by $s_1$ and the
  one described by $s_2$. The world $s$ in the state $(M,s)$
  identifies the world in $M[S]$ that corresponds to the actual world.
  
We will often view a Kripke structure $M=\langle S,\pi, \RK_{1},\ldots,\RK_{n} \rangle$ as a directed labeled graph,
whose set of nodes is $S$ and whose set of
edges contains $(s,i,t)$\footnote{$(s,i,t)$ denotes the edge from node $s$ to node $t$, labeled by $i$.} if and only if $(s,t) \in \RK_{i}$. $(s,i,t)$ is referred to as an edge coming out of (resp.\ into) the
   world $s$ (resp.\ $t$).

  Following \cite{vanDitmarschHK07}, we will refer to a pointed Kripke
  structure $(M,s)$ as a {\em state} and often use these two terms 
  interchangeably.

  The satisfaction relation between belief formulae and a state
is defined next. 

\begin{definition} 
\label{entailment-ml}
{
Given a formula $\varphi$, a Kripke structure 
$M = \langle S,\pi,\RK_{1},\ldots,\RK_{n} \rangle$, and a world $s \in S$:  
\begin{itemize} 
\item[(i)] $(M,s) \models \varphi$ if $p$ is a 
fluent and $M[\pi](s) \models p$; 

\item[(ii)] $(M,s) \models \B_{i}\varphi$ if for each $t$ such 
that $(s,t) \in \RK_{i}$, $(M,t) \models \varphi$;

\item[(iii)] $(M,s) \models \neg\varphi$ if $(M,s) \not\models \varphi$;

\item[(iv)] $(M,s) \models \varphi_1 \vee \varphi_2$ if $(M,s) \models \varphi_1$ or $(M,s) \models \varphi_2$;

\item[(v)] $(M,s) \models \varphi_1 \wedge \varphi_2$ if $(M,s)\models \varphi_1$ and $(M,s) \models \varphi_2$.

\item[(vi)] $(M,s) \models \mathbf{E}_\alpha \varphi$ if 
$(M,s)\models \B_i \varphi$ for every $i \in \alpha$.

\item[(vii)] $(M,s) \models \mathbf{C}_\alpha \varphi$ if 
$(M,s)\models \mathbf{E}^k_\alpha \varphi$ for every $k \ge 0$, 
where
	\begin{itemize}
		\item  $\mathbf{E}^0_\alpha \varphi =  \varphi$ and
		\item 	$\mathbf{E}^{k+1}_\alpha = \mathbf{E}_\alpha (\mathbf{E}^k_\alpha \varphi)$.
	\end{itemize} 
\end{itemize} 
}
\end{definition} 
For a Kripke structure $M$ and a formula $\varphi$,
$M\models \varphi$ denotes the fact that $(M,s) \models \varphi$ for
  each $s\in M[S]$.   
 
 $\models \varphi$ denotes the fact that $M\models \varphi$ for every Kripke
  structure $M$.

  \begin{example}
  [State]
  \label{ex2}
{ 
  Let us consider a simplified version of Example \ref{ex1} in which the agents 
  are concerned only with the status of the coin. 
  The three agents $A, B, C$ do not know 
  whether the coin has `heads' or `tails' up and this is a common belief. 
  Let us assume that the coin is heads up. 
  The beliefs of the agents about the world and about the beliefs of other agents 
  can be captured by the state of 
  Figure~\ref{htail}.
  
  \begin{figure}[htbp]
  \centerline{\includegraphics[width=.35\textwidth]{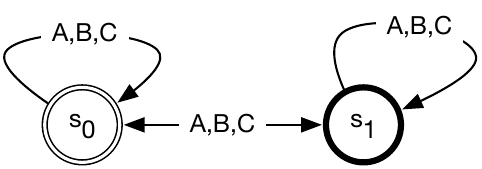}}
  \caption{An example of a state}
  \label{htail}
  \end{figure}
  
%
%
%
%
%
%
%
%
%
  
  In the  figure, a circle represents a world. 
  The name  of the world are 
  written in the circle. Labeled edges between worlds 
  denote the belief relations of the structure.  
  A double circle identifies the real world. 
  The interpretation of the world will be given whenever it is necessary. For example, 
  we write $M[\pi](s_0) = \{head\}$ to denote that $head$ is true in the world $s_0$ and 
  anything else is false. Similarly, $M[\pi](s_1) = \{ \}$ to denote that every fluent is false in the world $s_1$.
  }
  \end{example}

  We will occasionally be interested in Kripke structures that satisfy certain conditions. In particular, given
   a Kripke structure $M=\langle S, \pi, \RK_1, \dots, \RK_n\rangle$ we identify the following properties:
  \begin{itemize} 
  \item {\bf K}: for each agent $i$ and formulae $\varphi, \psi$, we have that 
  	$M \models (\B_i\varphi \wedge \B_i(\varphi \Rightarrow \psi)) \Rightarrow \B_i \psi$.
  \item {\bf T}: for each agent $i$ and formula $\psi$, we have that $M\models \B_i \psi \Rightarrow \psi$.
  \item {\bf 4}: for each agent $i$ and formula $\psi$, we have that
  	$M\models\B_i\psi \Rightarrow \B_i \B_i \psi$.
\item {\bf 5}: for each agent $i$ and formula $\psi$, we have that
	$M\models\neg \B_i \psi \Rightarrow \B_i \neg \B_i \psi$.
\item {\bf D}: for each agent $i$ we have that $M\models\neg \B_i \:\bot$.
  \end{itemize} 
  A Kripke structure is said to be a {\bf T}-Kripke ({\bf 4}-Kripke,
  {\bf K}-Kripke, {\bf 5}-Krikpe, {\bf D}-Kripke, respectively) structure if it satisfies
  property {\bf T} ({\bf 4}, {\bf K}, {\bf 5}, {\bf D}, respectively).
  A Kripke structure is said to be an {\bf S5} structure 
  if it satisfies the properties {\bf K}, {\bf T}, {\bf 4}, and {\bf 5}. The
  {\bf S5} properties have been often used to capture the notion
  of knowledge.
  Consistency of a set of formulae is defined next. 
  
  \begin{definition} 
  A  set of belief formulae $X$ is said to be p-{\em satisfiable} (or p-{\em consistent}) for $p \in \{\mathbf{S5, K, T, 4, 5}\}$ 
   if there exists a $p$-Kripke structure $M$ and a world  $s \in M[S]$ such that $(M,s) \models \psi$ for every $\psi \in X$. In this case,
  $(M,s)$ is  referred to as a $p$-\emph{model} of X. 
  \end{definition} 
  Finally, let us introduce a notion of equivalence between states. 

\begin{definition} \label{def-equivalence}
A state $(M,s)$ is \emph{equivalent} to 
a state $(M',s')$ if 
$(M,s) \models \varphi$ iff $(M',s') \models \varphi$
for every formula $\varphi \in {\cal L}_{\calag}$.
\end{definition} 

\subsection{Update Models}
\label{sub-update}

The formalism of \emph{update models} has been used to describe transformations of (pointed) Kripke structures
according to a predetermined transformation pattern. An update model is structured similarly 
to 
a pointed Kripke structure and it describes how to transform a pointed Kripke
structure using an update operator  defined in \citep{BaltagM04,BenthemEK06}. 

Let us start with some preliminary definitions.
An $\call_\calag$-substitution is a set  
$
\{p_1 \rightarrow \varphi_1,\ldots, p_k \rightarrow \varphi_k\},
$
where each $p_i$ is a distinct fluent in $\fluents$ 
and each $\varphi_i \in \call_\calag$. 
$SUB_{\call_\calag}$ denotes 
the set of all $\call_\calag$-substitutions.
\begin{definition}[Update Model]
\label{def-am}
Given a set $\calag$ of $n$ agents,
an {\em update model} $\mathbf{\Sigma}$ 
is a tuple $\langle\Sigma, R_1, \dots, R_n, pre, sub\rangle$ where 
\begin{itemize}
\item[(i)] 
$\Sigma$ is a set, whose elements are called \emph{events};
\item[(ii)] each $R_{i}$ is a binary relation on $\Sigma$;
\item[(iii)] $pre: \Sigma \rightarrow \call_\calag$ is a function mapping 
    each event $e \in \Sigma$ to a  formula in $\call_\calag$; and 
\item[(iv)] $sub: \Sigma \rightarrow \sub$ is a function mapping 
    each event $e \in \Sigma$ to a substitution in  $SUB_{\call_\calag}$. 
\end{itemize}
An \emph{update instance} $\omega$ is a pair $(\mathbf{\Sigma}, e)$ where 
$\mathbf{\Sigma}$ is an update model $\langle\Sigma, R_1,\dots, R_n, pre, sub\rangle$ 
and $e$, referred to as a {\em designated event}, is a member of $\Sigma$. 
\end{definition}
Intuitively, an update model represents different views of an action occurrence 
which are associated with the observability of agents. 
Each view is represented by an event in $\Sigma$. 
The designated event is the  one that agents who are aware of 
the action occurrence  will observe. 
The relation $R_i$ describes agent $i$'s uncertainty on action execution---i.e.,
if $(\sigma,\tau) \in R_i$ and event $\sigma$ is performed, then agent $i$
may believe that event $\tau$ is executed instead.
$pre$ defines the action precondition and $sub$ specifies the changes of 
fluent values after the execution of an action.

\begin{definition}[Updates by an Update Model]
\label{def:updo}
Let $M$ be a Kripke structure and 
$\mathbf{\Sigma} = \langle\Sigma, R_1,\dots,R_n, pre, sub\rangle$ be 
an update model. 
The \emph{update} induced by $\mathbf{\Sigma}$
defines a Kripke structure $M'=M \otimes \mathbf{\Sigma}$, where: 
\begin{itemize}
\item[({\em i})]
$M'[S] = \{(s,\tau) \mid s \in M[S], \tau \in \Sigma, (M,s) \models pre(\tau)\}$;
\item[({\em ii})]
$((s,\tau), (s',\tau')) \in M'[i]$ iff $(s,\tau),(s',\tau')\in M'[S]$, $(s,s') \in M[i]$ and $(\tau,\tau') \in R_{i}$;  
\item[({\em iii})] For all $(s,\tau) \in M'[S]$ and $f \in  \fluents$, $M'[\pi]((s,\tau))\models f$  iff $f \rightarrow \varphi \in  sub(\tau)$ 
     and $(M,s) {\models} \varphi$. 
\end{itemize}
\end{definition}
The structure $M'$ is obtained from the component-wise 
cross-product of the old structure $M$ and the 
update model $\mathbf{\Sigma}$, 
by {\bf ({i})} removing 
pairs $(s,\tau)$ such that $(M,s)$ does not satisfy the action precondition 
(checking for satisfaction of action's precondition), and {\bf({ii})}
removing links of the form $((s,\tau), (s',\tau'))$ 
from the cross product of $M[i]$ and $R_i$ 
if $(s,s') \not\in M[i]$ or $(\tau,\tau') \not\in R_i$ 
(ensuring that each agent's accessibility relation is updated according to  the update model). 

An {\em update template} is a pair $(\mathbf{\Sigma},\Gamma)$, where 
$\mathbf{\Sigma}$ is an update model with the set of events $\Sigma$ 
and $\Gamma \subseteq \Sigma$. 
The update of a state $(M,s)$ given an update template $(\mathbf{\Sigma},\Gamma)$ 
is a set of states, denoted by $(M,s) \otimes (\mathbf{\Sigma},\Gamma)$, 
where 
\[
(M,s) \otimes (\mathbf{\Sigma},\Gamma) 
 = \{(M \otimes \mathbf{\Sigma}, (s,\tau)) \mid \tau \in \Gamma, (M,s) \models pre(\tau)\}
\]


\begin{remark}
In the following, we will often represent an update instance by a graph 
with rectangles, double rectangles, and labeled links between rectangles 
representing events, designated events, and the relation of agents, respectively, 
as in the graphical representation of a Kripke structure. 
\end{remark} 

\subsection{Finitary {\bf S5}-Theories} 

A finitary {\bf S5}-theory, introduced in \citep{SonPBG14}, is a collection of formulae 
which has finitely many {\bf S5}-models, up to a notion 
of equivalence (Definition~\ref{def-equivalence}). To define finitary {\bf S5}-theories, we 
need the following notion.  Given a set of propositions
$\fluents$,  a \emph{complete clause} over $\fluents$ is  a disjunction of the form 
$\bigvee_{p \in \fluents} p^*$ where $p^*$ is either $p$ or $\neg p$.  
We will consider formulae of the following forms:
\begin{align} 
  & \varphi & \label{init1}\\
   & \mathbf{C} (\B_i \varphi) &  \label{init21} \\
   & \mathbf{C} (\B_i \varphi \vee \B_i \neg \varphi) & \label{init3}\\
   & \mathbf{C} (\neg \B_i \varphi \wedge \neg \B_i \neg \varphi) & \label{init4}
\end{align}
where $\varphi$ is a fluent formula.
\begin{definition} 
\label{def:finitary}
A theory $T$ is said to be {\em primitive finitary {\bf S5}} if 
\begin{itemize} 
\item Each formula in $T$ is of the form (\ref{init1})-(\ref{init4}); and 
\item For each complete clause $\varphi$ over $\fluents$ 
and each agent $i$, $T$ contains either
 {\bf (i)} $\mathbf{C}(\B_i \varphi)$ or 
 {\bf (ii)} $\mathbf{C}(\B_i \varphi \vee \B_i \neg \varphi)$ or
 {\bf (iii)}  $\mathbf{C}(\neg \B_i \varphi \wedge \neg \B_i \neg \varphi)$.
\end{itemize} 

A  theory $T$ is a {\em finitary {\bf S5}-theory} if 
$T \models H$ and $H$ is a primitive finitary {\bf S5}-theory. 

$T$ is {\em pure} if $T$ contains only formulae of the form \eqref{init1}-\eqref{init4}.
\end{definition} 

We say that a state $(M,s)$ is  {\em canonical} if for every pairs of worlds 
 $u, v\in M[S]$ and $u \ne v $, $M[\pi](u) \not\equiv M[\pi](v)$ holds. We have that 

\begin{theorem} 
[From \citep{SonPBG14}]
\label{theorem:finitary}  
Every finitary {\bf S5}-theory $T$ has finitely many finite canonical models, up to equivalence. 
If $T$ is pure then these models are minimal and their structures are identical up to the name 
of the points. 
\end{theorem} 

\subsection{Why an Action Language?} 


As  mentioned earlier, the Dagstuhl seminars \citep{ALLB14,BBMvD17} identified one of the main research topics in EPM: \emph{the 
development of an adequate specification language for EPM}. This problem arises from the fact that EPM has been defined and investigated using a DEL based approach, in which actions are represented by update models  (Subsection~\ref{sub-update}). This representation has been useful for the understanding of EPM and the study of its complexity, but  comes with a significant drawback---practical and conceptual knowledge representation challenges---discussed  by Herzig\footnote{\tiny \url{http://materials.dagstuhl.de/files/17/17231/17231.AndreasHerzig.Slides.pdf}} at the   Dagstuhl seminar \cite{BBMvD17}. 
Let us consider a slight modification of Example~\ref{ex1}, where the box is open,  $A$ has looked at the coin, 
while  both $B$ and $C$ are distracted, and  $A$ can announce whether the coin lies heads up or tails up. However, only agents who are attentive (to $A$) could listen to what $A$ says. Assume that $A$ announces that the coin lies heads up. Intuitively, this action occurrence can have different effects on the beliefs of the other agents---depending on the context and the specific features of each of them, e.g., whether  the agent is attentive to $A$. As a result, we need a variety of  update models to represent this primitive action. Herzig refers to this problem  as the 
 \emph{action type vs. action token} problem. 

\begin{figure}[h]
\centering{\includegraphics[width=\textwidth]{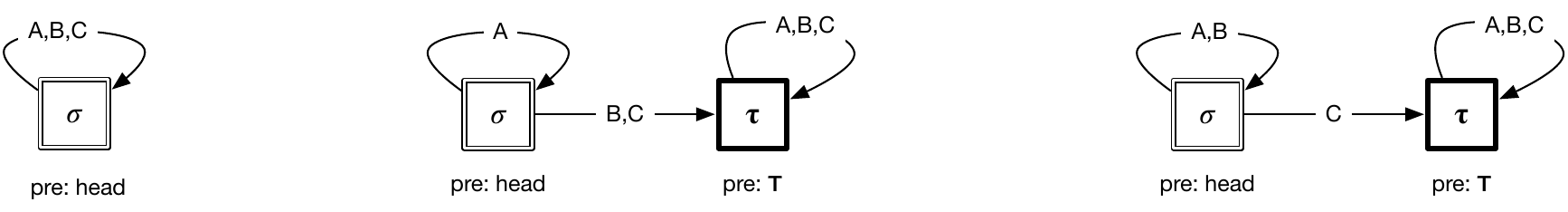}} 
\caption{\small Update models for announcing \emph{``the coin lies heads up''} by $A$ in different situations} 
\label{fig:announcing_coin} 
\end{figure} 

Fig.~\ref{fig:announcing_coin} shows three update models; they describe
 the occurrence of the announcement by $A$, stating that the coin lies heads up, assuming that the coin indeed lies heads up in the real state of the world. On the left is the update model when both $B$ and $C$ are attentive.
 The model in the middle depicts the situation when both $B$ nor $C$ are not attentive.
 The update model on the right captures the case of $B$ being attentive and $C$ being not attentive. 
 In the figures, $\sigma$ and $\tau$ are events and $\sigma$ is a
\emph{designated event,} $head$ is a propositional variable denoting that the coin lies heads up. 

Observe that these models are used only when the coin indeed lies heads up. The update models corresponding to the situation
 where $head$ is false (i.e., when $A$ makes a false announcement) in the real state of the world are different from those in the figure and are omitted. It is easy to see that  the number of update models needed to represent such simple announcement of  \emph{``the coin lies heads up''} by $A$ is exponential in the number of 
agents. This is certainly an undesirable consequence of using update models and epistemic actions for representing and reasoning about effects of actions in multi-agent domains. Therefore, any specification language for representing and reasoning about the effects of actions in multi-agent domains should consider that the announcement of the coin lies heads up by $A$ is simply a \emph{primitive action}. In our view, the 
update models should be derived from the concrete state, which is a combination of the real state of the world and the state of beliefs of the agents, and not specified directly. 
A more detailed discussion on this issue can be found in Section~\ref{related}, the related work section. 



\section{The language \mastar: Syntax} 
\label{syntax} 

In this paper, we consider multi-agent domains in which 
the agents are truthful and no false information may be announced
or observed. Furthermore, the underlying 
assumptions guiding the semantics of our language are the rationality
principle and the idea that beliefs of an agent are inertial.
In other words, agents  believe something because they have a reason
to, and the beliefs of 
an agent remain the same unless something causes them to change.

In this section and in the next section, we introduce 
the language \mastar{} for describing 
actions and their effects in multi-agent environment. 
The language is built over a signature $\langle \agents, \calf, \cala\rangle$,
where $\agents$ is a finite set of agent identifiers, 
$\calf$ is a set of fluents, and $\cala$ is a set of actions. 
Each action in $\cala$ is an action the agents 
in the domain are capable of performing.  

Similar to any action language developed for single-agent environments, 
\mastar{} consists of three components which will be used in describing 
the actions and their effects, the initial state, and the query language
(see, e.g., \citep{GelfondL98}). 
We will next present each of these components. Before we do so, let us 
denote the multi-agent domain in Example~\ref{ex1} by $D_1$. 
For this domain, we have that $\calag = \{A, B, C\}$.  The set of 
fluents $\calf$ for this domain consists of: 
\begin{itemize}
\item $tail$: the coin lies tails up ($head$ is often used in place of $\neg tail$); 
\item $has\_key(x)$: agent $x$ has the key of the box;
\item $opened$: the box is open; and 
\item $looking(x)$: agent $x$ is looking at the box. 
\end{itemize}
The set of actions for $D_1$ consists of: 
\begin{itemize}
\item $open$: an agent opens the box; 
\item $peek$: an agent peeks into the box; 
\item $signal(y)$: an agent signals agent $y$ (to look at the box); 
\item $distract(y)$: an agent distracts agent $y$ (so that $y$ does not 
look at the box); and  
\item $shout\_tail$: an agent  announces that the coin lies tails up. 
\end{itemize}
where $x, y \in \{A, B, C\}$. We start with the description of actions 
and their effects.

\subsection{Actions and effects}

We envision three types of actions that an 
agent can perform: \emph{world-altering actions} (also known as {\em ontic actions}), 
\emph{sensing actions}, and \emph{announcement actions}. Intuitively,
\begin{itemize}
\item A world-altering action is used to explicitly modify certain properties of the 
       world---e.g., the agent $A$ opens the box 
        in Example \ref{ex1}, or the
        agent $A$ distracts the agent $B$ so that $B$ does not look at the box 
        (also in Example \ref{ex1});
\item A sensing action is used by an agent to refine its beliefs about the world, by making direct 
	observations---e.g., an agent peeks into the box; the effect
	of the sensing action is to reduce the amount of uncertainty of the agent;
\item An announcement action is used by an agent to affect the beliefs of the agents receiving the communication---we
	operate under the assumption that agents receiving an announcement always believe what is being announced.
\end{itemize} 
For the sake of simplicity, we assume that each action $a\in \cala$ falls in exactly 
one of the three categories.\footnote{It is
easy to relax this condition, but it would make the presentation more tedious.} 
 In a multi-agent system, we need
to identify an action occurrence with the agents who execute it. Given
an action $a \in \cala$
and a set of agents $\alpha \subseteq \calag$, we write ${\sf a}\langle \alpha\rangle$ to denote the joint-execution of
$a$ by the agents in $\alpha$ and call it an \emph{action instance}. 
We will use $\calai$ to denote the set of possible action instances $\cala \times 2^\calag$. 
Elements of $\calai$ will be written in sans-serif font. 
For simplicity of the presentation, we often use ``an action'' or ``an action instance'' interchangeably when it is clear from the context 
which term is appropriate.  
Furthermore, when $\alpha$ is a singleton set, $\{x\}$, we simplify $\langle \{x\} \rangle$ to $\langle x \rangle$. 


In general, an action can be executed only under certain conditions, 
called its {\em executability conditions}.
For example, the statement ``to open the box, an agent must have its key'' in Example~\ref{ex1} describes the executability condition of 
the action of opening a box.  
The first type of statements  in \mastar{} is used to describe the executability conditions of action occurrences and is of the following form:

\begin{equation} \label{exec}
 \executable {\sf a} \iif \psi  
\end{equation}
where ${\sf a} \in \calai$  and $\psi$ is a belief formula. A statement of 
type \eqref{exec} will be referred to as the {\em executability 
condition of the action occurrence ${\sf a}$}. $\psi$ is referred as the {\em precondition of} ${\sf a}$.
For simplicity of the presentation, we will assume that each action occurrence 
${\sf a}$ is associated with exactly one executability condition. When $\psi = \top$, the statement will be omitted. 
 
For an occurrence of a world-altering action ${\sf a}$, such as the action of opening the box by some agent,  
we have statements of the following type that express the 
change that may be caused by such action:
\begin{equation} \label{causes}
{\sf a} \causes \ell \iif \psi
\end{equation}
where $\ell$ is a fluent literal and $\psi$ is a belief formula. 
Intuitively, if the real state of the world and the beliefs match the condition 
described by $\psi$, then the real state of the world is affected by the change
that makes the literal $\ell$ true
 after the execution of $a$. When $\psi = \top$, the part 
``$\iif \psi$'' will be omitted from \eqref{causes}. 
We also use  
$$
{\sf a} \causes \phi \iif \psi
$$
where $\phi$ is a set of fluent literals as a shorthand for 
the set $\{{\sf a} \causes \ell \iif \psi \mid \ell \in \phi\}$.

Sensing actions, such as the action of looking into the box, 
allow agents to learn about the value of a fluent in the real 
state of the world (e.g., learn whether the coin lies heads or tails up).  We use 
statements of the following kind to represent effects of sensing action occurrences:
\begin{equation} \label{sense}
 {\sf a}  \determines \varphi  
\end{equation}
where $\varphi$ is a fluent formula   and ${\sf a} \in \calai$ is a sensing action. 
Statements of type \eqref{sense} encode the occurrence of 
a sensing action ${\sf a}$ which enables the agent(s) to learn  the value 
of the fluent formula $\varphi$. $\varphi$ is referred to as a {\em sensed fluent formula} of ${\sf a}$.  

For actions such as the action of an agent telling
another agent that the coin lies heads up,  
we have statements of the following kind:

\begin{equation} \label{announce}
{\sf a}  \announces \varphi 
\end{equation}
where $\varphi$ is a fluent formula and ${\sf a} \in \calai$. 
${\sf a}$ is called {\em an announcement action}. 

We will next illustrate the use of statements of the form \eqref{exec}-\eqref{announce} 
to represent the actions of the domain $D_1$. 

\begin{example}
\label{ex4} 
{\rm 
The actions of domain $D_1$ can be specified by the following statements: 
\[
\begin{array}{l} 
   \executable  open\langle x \rangle \iif has\_key(x)    \\
   \executable  peek\langle x \rangle \iif opened, looking(x)    \\  
   \executable  shout\_tail \langle x \rangle \iif \B_x(tail), tail \\
   \executable signal(y)\langle x \rangle \iif looking(x), \neg looking(y) \\ 
   \executable distract(y)\langle x \rangle \iif looking(x), looking(y) \\ 
   \\
   open  \langle x \rangle \causes opened \\ 
   signal(y)\langle x \rangle \causes looking(y) \\
   distract(y)\langle x \rangle \causes \neg looking(y) \\ 
   peek\langle x \rangle \determines tail \\ 
   shout\_tail \langle x \rangle \announces tail \\ 
\end{array}
\]
where $x$ and $y$ are different agents in $\{A, B, C\}$. The first five statements 
encode the executability conditions of the five actions in the domain. 
The next three statements describe the effects of the occurrence of three world-altering actions. 
$peek \langle x \rangle$ is an example of an instance of a sensing action. Finally, $shout\_tail \langle x \rangle$ 
is an example of an instance of an announcement action. 
}
\end{example} 

\subsection{Observability: observers, partial observers, and others}

One of the key differences between single-agent and multi-agent domains lies
in how the execution of an action changes the beliefs of agents. 
This is because, in multi-agent domains, an agent might be oblivious 
about the occurrence of an action or unable to observe the effects of an 
action. For example, watching another agent open the box 
would allow the agent to know that the box is open after the execution of the action; 
however, the agent would still believe that the box is closed if she is not
aware of the action occurrence. On the other hand, watching   
another agent peek into the box does not help the observer in 
learning whether the coin lies heads or tails up; the only thing
 she would  learn is that 
the agent who is peeking into the box has  knowledge of the status of the coin.  

\mastar{} needs to  have a component for representing    
the fact that not all the agents may be completely aware of the occurrence of
actions being executed. Depending on the action and the current situation, we can 
categorize agents in three classes: 
\begin{itemize}
\item \emph{Full observers}, 
\item \emph{Partial observers}, and 
\item \emph{Oblivious} (or \emph{others}).
\end{itemize} 
 This categorization is dynamic---changes in the state of the world
 and/or the beliefs of agents  may change the observability of actions. 
 In this paper, we will consider the 
 possible  observabilities of agents for different action types as
 detailed in Table~\ref{roles}. 

\begin{table}[htbp]
\begin{center}
\begin{tabular}{|l|c|c|c|}
  \hline\hline
 \multicolumn{1}{|c|}{\bf   action type} & {\bf full observers} & {\bf partial observers} & {\bf oblivious/others}\\
 \hline
 \hline 
  \emph{world-altering actions}          & $\surd$ &   &   $\surd$      \\ %
  \emph{sensing actions} &   $\surd$ & $\surd$ & $\surd$ \\
  \emph{announcement actions} & $\surd$ & $\surd$ & $\surd$  \\
  \hline
\end{tabular}
\caption{Action types and agent observability}\label{roles}
\end{center}
\end{table}

The first row indicates that, for a world-altering action, an agent can either be a \emph{full observer},
i.e., completely aware of the occurrence of that action,  or {oblivious} of the occurrence
 of the action. The assumption here is that agents are fully knowledgeable of the outcomes of a
 world-changing action (which makes partial observability a moot point). 
 In the second  case,  the observability of the agent is categorized as \emph{other}. 
 Note that we assume that the observer agents know about each others' status and 
 they are also aware of the fact that the other agents are oblivious. The oblivious agents have no clue of anything.
 Notice also that in multi-agent domain, an agent, who executes an action, might not be a full observer of the action occurrence. 
For instance, a blind agent will not be able to observe the effect of switching the contact of a light. 

For a sensing action, an agent can either be a
\emph{full observer}, i.e., it is aware of the occurrence of that action and of its results,
it can be a \emph{partial observer}, gaining knowledge that the full observers have 
performed a sensing action  but without knowledge
of the result of the observation,   or it can be oblivious of the occurrence of the action (i.e., 
\emph{other}). 
Once again, we assume that the observer agents know about each others' status and they
also know about the agents partially observing the action and about the agents that are oblivious. 
The partially observing agents know about each others' status, and they also know about the 
observing agents and the agents that are oblivious. The oblivious agents have no clue of anything.
The behavior is analogous for the case of announcement actions.

We observe that this classification is limited to individual agent's observability of action occurrences. 
As agents are knowledgeable about the domains, they can reason about others' observability when an action occurs and manipulate others' observability, thereby others' knowledge about the world and beliefs. As such, it is reasonable for an agent to use \mastar{} for planning purpose, e.g., in Example~\ref{ex1}, $A$ distracts $C$ and signals $B$ before opens the box. The classification, however, does not consider situations in which an agent might have uncertainty about others' observability. We discuss this limitation in Sub-subsection~\ref{subsub:expressivity}. A possible way to address this issue is to remove the assumption that agents are oblivious of actions' occurrences by default. This topic of research is interesting in its own right and deserves a throughout investigation. We leave this as a future work.  

Agents' observability is meant to be dynamic and the dynamic behavior is described by
 agent observability  
statements of the following forms:\footnote{As discussed earlier, the ``$\iif \top$'' are omitted
from the statements.} 
\begin{eqnarray} 
 & z  \observes {\sf a}   \iif \varphi \label{observes}\\
 & z \pobserves {\sf a}  \iif \psi \label{pobserves}
\end{eqnarray}
where $z \in \agents$, ${\sf a} \in \calai$, and 
$\varphi$ and $\psi$ are fluent formulae. 
\eqref{observes} indicates that agent $z$ is a 
full observer of   ${\sf a} $ if $\varphi$ holds. 
\eqref{pobserves} states that agent $z$ is a   
partial observer of   ${\sf a} $ if $\psi$ holds.
$z$, ${\sf a}$, and $\varphi$ (resp. $\psi$) are referred to as the observed agent, 
the action instance, and the condition of \eqref{observes} (resp. \eqref{pobserves}). 
The next example illustrates the use of the above statements in specifying the agents 
observability  of the domain $D_1$. 

\begin{example} 
[Observability in $D_1$] 
\label{ex5} 
{
The actions of $D_1$ are described in Example~\ref{ex4}. 
The observability of agents in $D_1$ can be described by the set $O_1$ of statements 
\[
\begin{array}{lcl} 
x \observes open\langle x \rangle  & \hspace{.6cm} & x \observes peek\langle x \rangle \\
y \observes open\langle x \rangle  \iif looking(y)  && y \pobserves peek\langle x \rangle  \iif looking(y) \\
z \observes shout\_tail \langle x \rangle  && \forall z\in \{A,B,C\}\\
x \observes distract(y)\langle x \rangle  && x \observes signal(y)\langle x \rangle \\
y \observes distract(y)\langle x \rangle  && y \observes signal(y)\langle x \rangle \\
\end{array} 
\]
where  $x$ and $y$ denote different agents in $\{A,B,C\}$.   
The above statements say that agent $x$ is a 
fully observant agent when $open\langle x \rangle$, $peek\langle x \rangle$, 
$distract(y)\langle x \rangle$, $signal(y)\langle x \rangle$, or $shout\_tail\langle x \rangle$ 
are executed; $y$ is a fully observant agent if it is looking 
(at the box) when $open\langle x \rangle$ is executed.  
$y$ is a partially observant agent if it is looking 
when $peek\langle x \rangle$ is executed. An agent different from $x$ and $y$  
is oblivious in all cases. 
} 
\end{example} 

It is obvious that an agent cannot be both partially observable and fully observation of the execution of an action at the same time. For this reason, 
we will say that a statement of the form \eqref{observes} is in conflict with 
a statement of the form \eqref{pobserves} if  for the same occurrence ${\sf a} \in \calai$ and $z\in \agents$,  $\varphi \wedge \psi$ is consistent.

%
%
\begin{definition} 
An \mastar{} \emph{domain} is a collection of statements of
the forms (\ref{exec})-(\ref{pobserves}). 
\end{definition} 
Similarly to action domains in the language $\cal A$ introduced by \cite{GelfondL93}, an \mastar{} domain could contain two statements specifying contradictory effects of an action occurrence such as 
\[
{\sf a} \causes f \iif \varphi \quad \quad \textnormal{and} \quad \quad 
{\sf a} \causes \neg f \iif \psi  
\]
where $\varphi \wedge \psi$ is a consistent formula, i.e., there exists some pointed Kripke structure $(M,s)$ such that 
$(M,s) \models \varphi \wedge \psi$. Such a domain is not sensible and will be characterized as \emph{inconsistent}. 
\begin{definition} 
An \mastar{} \emph{domain} $D$ is \emph{consistent} if for every pointed Kripke structure $(M,s)$ and 

\begin{itemize} 
\item for every pair of 
two statements 
\[
{\sf a} \causes f \iif \varphi \quad \quad \textnormal{and} \quad \quad 
{\sf a} \causes \neg f \iif \psi  
\]
in $D$,  $(M,s) \not\models \varphi \wedge \psi$; and 

\item for  every pair of  
two statements 
\[
z \observes {\sf a}  \iif \varphi \quad \quad \textnormal{and} \quad \quad 
z \pobserves {\sf a} \iif \psi  
\]
in $D$,  $(M,s) \not\models \varphi \wedge \psi$.
\end{itemize} 
\end{definition} 
\noindent From now on, whenever we say an \mastar{} domain $D$, we will assume that $D$ is consistent.

\subsection{Initial State}

A domain specification encodes the actions, and their effects, 
and the observability of agents in each situation. 
The initial state, that encodes both the initial state of the world 
and the initial beliefs of the agents, is specified in \mastar{} 
using \emph{initial statements} of the following form:

\begin{equation} \label{init}
\initially \varphi 
\end{equation}

\noindent 
where $\varphi$ is a belief formula. Intuitively, this statement says that 
$\varphi$ is true in the initial state. We will later 
discuss restrictions on the formula $\varphi$ to ensure the
computability of the Kripke structures describing the
initial state.

\begin{example} 
[Representing Initial State of $D_1$] 
\label{ex3}
{\rm
Let us reconsider Example \ref{ex1}. The initial state of $D_1$ 
can be expressed by the following 
statements: 
\[
\begin{array}{ll}  
\initially \mathbf{C}(  has\_key(A) )\\
\initially \mathbf{C}(\neg  has\_key(B) )\\
\initially \mathbf{C}(\neg  has\_key(C) )\\
\initially \mathbf{C}(  \neg opened )\\ 
\initially \mathbf{C}(  \neg \B_x head \wedge \neg \B_x \neg head ) & \:\:\textnormal{for } \: x \in \{A,B,C\}\\
\initially \mathbf{C}( looking(x))  & \:\:\textnormal{for } \: x \in \{A,B,C\}\\
\end{array}
\]
These statements indicate that everyone knows that $A$ has the key 
and $B$ and $C$ do not have the key, the box is closed, no one knows 
whether the coin lies heads or tails up, and everyone is looking at the box.  
}
\end{example} 
The notion of an action theory in \mastar{}
is defined next. 

\begin{definition} 
[Action Theory]
An \mastar{}-{\em action theory} is a pair $(I, D)$ where $D$ is an
\mastar{} domain and $I$ is a set of statements of the form \eqref{init}.
\end{definition}

In Section~\ref{semantics}, we will define the notion of entailment between action theories
and queries, in a manner similar to the notion of entailment defined for action languages in single-agent domains (e.g., \citep{GelfondL98}).
This requires the following definition. 

\begin{definition} 
[Initial State/Belief-State]
\label{def:init_state}
Let $(I, D)$ be an action theory. An {\em initial state} of $(I, D)$ is a 
state $(M,s)$ such that for every statement $$\initially \varphi$$ in 
$I$, $(M,s) \models \varphi$. 

$(M,s)$ is an initial {\bf S5}-state 
if it is an initial state and $M$ is a {\bf S5} Kripke structure. 

The {\em initial belief-state} (or \emph{initial b-state}) of $(I, D)$ is the collection of all 
initial states of $(I, D)$. 

The {\em initial {\bf S5}-belief state} of $(I, D)$ is the collection of all 
initial {\bf S5}-states of $(I, D)$. 
\end{definition}   


By definition, it is easy to see that, theoretically, 
there could be infinitely many initial states for an
arbitrary \mastar{} theory. For example,
given a state $(M,s)$ and a set of formulae $\Sigma$ 
such that $(M,s) \models \Sigma$, a new state
$(M',s)$ that also satisfies $\Sigma$ can be 
constructed from $M$ by simply stating $M'[S] = M[S] \cup \{s'\}$, where
$s'\not\in M[S]$, and keeping everything else unchanged. As such, it is important
to identify sensible 
classes of action theories whose initial belief states are finite, up to a notion 
of equivalence (Definition~\ref{def-equivalence}). Fortunately, the result on 
finitary {\bf S5} theories\footnote{
  To keep the paper at a reasonable length, we do not discuss the details of 
   finitary {\bf S5} theories. Interested readers are referred to \citep{SonPBG14} for details 
   and proof of Theorem~\ref{theorem:finitary}.
}  (Definition~\ref{def:finitary})  allows us to identify a large class
of action theories satisfying this property. We call them definite action theories and 
define them as follows. 

\begin{definition}
[Definite Action Theory] 
\label{definite-theory}
An action theory $(I,D)$ is said to be {\em definite} if  
the theory $\{\varphi \mid \initially \varphi$ belongs to $I\}$ 
is a finitary-{\bf S5} theory. 
\end{definition} 

Observe that Theorem~\ref{theorem:finitary} indicates that for definite action theories, 
the initial belief state is finite. An algorithm for computing the initial belief state 
is given in \citep{SonPBG14}. This, together with the definition of the transition function 
of \mastar{} domains in the next section, allows  the implementation of search-based
progression epistemic planning systems. A preliminary development can be found 
in \citep{LeFSP18}. 

It is worth pointing out that Definition~\ref{def:init_state} does not 
impose any condition on the initial state of action theories. It merely characterizes a subgroup 
of action theories ({\bf S5} action theories), for which some interesting properties can be proved.  
We would also like to reiterate that  most of our discussion in this paper focuses on beliefs 
rather than knowledge. Additional steps need to be taken for answering questions related 
to knowledge of agents after the execution of an action sequence. For example, the ability of 
maintaining the KD45 properties of the resulting pointed Kripke structures after the execution 
of an action will be important. Preliminary investigation in this direction was presented in 
\cite{SonPBG15}.

\section{Update Model Based Semantics for \mastar{} Domains} 
\label{sec:updatemodel}
\label{semantics}

An \mastar{} domain $D$ specifies a transition system, whose nodes are 
states. This transition system will be 
described by a transition function $\Phi_D$, which maps pairs of action occurrences 
and states to states. For 
simplicity of the presentation, we assume that only one action occurrence happens  
at each point in time---it is relatively simple to 
extend it to cases where concurrent actions are present, and this is
left as future work. As we have mentioned in Section~\ref{background}, we will 
use pointed Kripke structures to represent states in \mastar{} action theories. 
A pointed Kripke structure encodes  three 
components: 
\begin{itemize} 
\item The actual world; 
\item The state of beliefs of each agent about the real state of the world; and  
\item The state of beliefs of each agent about the beliefs of other agents. 
\end{itemize} 
These components are affected by the execution of actions. Observe that
the notion of a state in \mastar{} action theories is  more complex than
the notion of state used in single-agent domains (i.e., a complete
set of fluent literals).

Let $\mathcal{S}$ be the set of all possible pointed Kripke structures  over
$\mathcal{L}(\calf, \agents)$, the transition function $\Phi_D$ maps 
pairs of action instances and states into sets of states, i.e.,  
\begin{equation} 
\Phi_D : \calai \times \mathcal{S} \longrightarrow 2^\mathcal{S}
\end{equation} 
will be defined for each action type separately and in two steps. First, we
define an update model representing the occurrence of ${\sf a} \in \calai$ in a state  
$(M,s)$. Second, we use the update model/instance  
defined in step one to define $\Phi_D$.
We start by defining the notion 
of a frame of reference in order to define the function $\Phi_D$.

\subsection{Actions Visibility and Frames of Reference}
Given a state $(M,s)$ and an action occurrence ${\sf a}$, let us define

\[ 
\begin{array}{lcl}
  \textit{F}_D({\sf a} ,M,s) & = & \{ x \in \agents \:|\: [\textit{x} \observes {\sf a}  \iif \varphi] \in D \textnormal{ such that } (M,s)\models \varphi\}\\
  \textit{P}_D({\sf a} ,M,s) & = & \{ x \in \agents \:|\: [\textit{x} \pobserves {\sf a}  \iif \varphi] \in D \textnormal{ such that } (M,s)\models \varphi\}\\
  \textit{O}_D({\sf a} ,M,s) & = & \agents \setminus (\textit{F}_D({\sf a} ,M,s) \cup \textit{P}_D({\sf a} ,M,s))
  \end{array}
  \]

 We will refer to 
 the tuple $(\textit{F}_D({\sf a} ,M,s), \textit{P}_D({\sf a} ,M,s), \textit{O}_D({\sf a} ,M,s))$ as the
 \emph{frame of reference} for the execution of ${\sf a} $ in $(M,s)$. 
 Intuitively, {\em F}$_D({\sf a} ,M,s)$ (resp. $\textit{P}_D({\sf a} ,M,s)$ and $\textit{O}_D({\sf a} ,M,s)$) are the agents
that are fully observant (resp. partially observant and oblivious/other) 
of the execution of   ${\sf a} $ in the state $(M,s)$. As we assume that for each 
pair of an action occurrence ${\sf a}$ and a  state $(M,s)$, 
the sets $(\textit{F}_D({\sf a} ,M,s)$, $\textit{F}_D({\sf a} ,M,s)$, and  $\textit{P}_D({\sf a} ,M,s)$ are pairwise disjoint,  
the domain specification $D$ and the state $(M,s)$
determine a unique frame of reference for each action occurrence.

The introduction of frames of reference allows us to elegantly 
model several types of actions that are aimed at modifying the frame 
of reference (referred to as \emph{reference setting actions}). Some 
possibilities are  illustrated in the following examples.

\begin{example}
[Reference Setting Actions]
{
Example~\ref{ex5} shows two reference setting actions 
{\it signal}$(y)$ and {\it distract}$(y)$ with instances of the form 
{\it signal}$(y)\langle x \rangle $ and {\it distract}$(y)\langle x \rangle $ because 
they change the truth value of  $looking(y)$ to true and false, respectively,
which decides whether or not the agent $y$ is aware (or partially aware) of the occurrence of an action  
instance $open\langle x \rangle$ (or $peek \langle x \rangle$).

The action instance {\it signal}$(y)\langle x \rangle $ allows agent $x$ to promote 
agent $y$ into a higher level of observation 
for the effect of  {\it peek}$\langle x \rangle $.  
On the other hand, the action instance {\it distract}$(y)\langle x \rangle $ 
allows agent $x$ to demote agent $y$ into a lower level of observation.
The net effect of executing these actions is a change of frame. 

Let us consider  $\textit{signal}(y)\langle x \rangle $ and a state $(M,s)$. 
Furthermore, let us assume that $(M',s')$ is a state resulting from 
the execution of $\textit{signal}(y)\langle x \rangle $ in $(M,s)$. 
The frames of reference for the execution of the action instance ${\sf a} = peek\langle x \rangle $ in these 
two states are related to each other by the following equations:
\[
\begin{array}{lll} 
F_{D_1}({\sf a},M',s') & = &F_{D_1}({\sf a},M,s) \\
P_{D_1}({\sf a},M',s') & = &P_{D_1}({\sf a},M,s)  \cup \{y\}\\
O_{D_1}({\sf a},M',s') & = & O_{D_1}({\sf a},M,s) \setminus \{y\} 
\end{array} 
\]

%

\noindent
Intuitively, after the execution of $signal(y)\langle x \rangle $, $looking(y)$ becomes true 
because of the statement $$signal(y)\langle x \rangle  \causes looking(y)$$ in $D_1$.
By definition, the statement $$y \pobserves peek\langle x \rangle  \iif looking(y)$$ indicates
that $y$ is partially observant. 

Similar argument shows that $distract(y)\langle x \rangle$ demotes $y$ to the
lowest level of visibility, i.e., it
will cause agent $y$ to become oblivious of the successive {\it peek}$\langle x \rangle$ action. 
Let us assume that the execution of ${\sf a}$ 
in $(M,s)$ resulted in $(M',s')$. Then, 
the frames of reference for the execution of the action instance ${\sf a} = peek\langle x \rangle $ in these 
two states are related to each other by the following equations:
\[
\begin{array}{lll} 
F_{D_1}({\sf a},M',s') & = &F_{D_1}({\sf a},M,s) \setminus \{y\}  \\
P_{D_1}({\sf a},M',s') & = &P_{D_1}({\sf a},M,s)  \setminus \{y\} \\
O_{D_1}({\sf a},M',s') & = & O_{D_1}({\sf a},M,s) \cup \{y\}
\end{array} 
\]
%
}
\end{example}

\subsection{Update Model for Action Occurrences} 
  
\begin{definition}[Update Model/Instance for World-Altering Actions] 
\label{upd-wa}
Given a world-altering action instance ${\sf a}$ with the precondition $\psi$
and a frame of reference 
$\rho=(F,\emptyset,O)$, the update model for ${\sf a}$ and $\rho$, 
denoted by $\omega({\sf a},\rho)$, 
is defined by  $\langle \Sigma, R_1,\dots, R_n,pre,sub\rangle$ where
\begin{itemize}
\item[$\circ$] $\Sigma = \{\sigma, \epsilon\}$;
\item[$\circ$] $R_i = \{(\sigma,\sigma),(\epsilon,\epsilon)\}$ for $i \in F$ and 
	$R_i = \{(\sigma,\epsilon),(\epsilon,\epsilon)\}$ for $i \in O$;
\item[$\circ$] $pre(\sigma) = \psi$ and $pre(\epsilon)=\top$; and 
\item[$\circ$] $sub(\epsilon) = \emptyset$ and 
$sub(\sigma) = \{p \rightarrow \Psi^+(p,{\sf a}) \vee (p \wedge \neg \Psi^-(p,{\sf a})) \mid 
                     p \in \fluents\},$
where \\
$$\Psi^+(p,{\sf a}) = \bigvee \{ \varphi\:|\: [ {\sf a} \causes p \iif \varphi ] \in D \}$$ 
and 
$$\Psi^-(p,{\sf a}) = \bigvee \{\varphi \:|\: [ {\sf a} \causes \neg p \iif \varphi ] \in D \}.$$  
\end{itemize}
The update instance for the occurrence of ${\sf a}$ and the frame of reference $\rho$ is $(\omega({\sf a},\rho), \{\sigma\})$.
\end{definition}
Observe that the update model of the world-altering action occurrence  has two 
events. Each event is associated to a group of agents in the frame of reference. 
The links in the update model 
for each group of agents reflect the state of beliefs each group would have 
after the execution of the action. For example, fully observant agents
(in $F$) will have no uncertainty.
The next example illustrates this definition.

\begin{example}
{ 
Going back to our original example, consider the occurrence of the action instance $open\langle A \rangle$  assuming that 
everyone is aware that $C$ is not looking at the box while $B$ and $A$ are. 
Figure \ref{umopen} (left) depicts the state $(M,s_0)$ where 
$M[\pi](s_0)=\{looking(A),looking(B),has\_key(A)\}$
and $M[\pi](s_1)=\{looking(A),looking(B),has\_key(A),head\}$. 
The frame of reference for $open\langle A \rangle$ in this situation 
is $(\{A,B\},\emptyset,\{C\})$.  The corresponding update
instance for $open\langle A \rangle$ and the frame of reference $(\{A,B\},\emptyset,\{C\})$ 
is given in Figure \ref{umopen} (middle). 

\begin{figure}[htpb]
\centerline{\includegraphics[width=\textwidth]{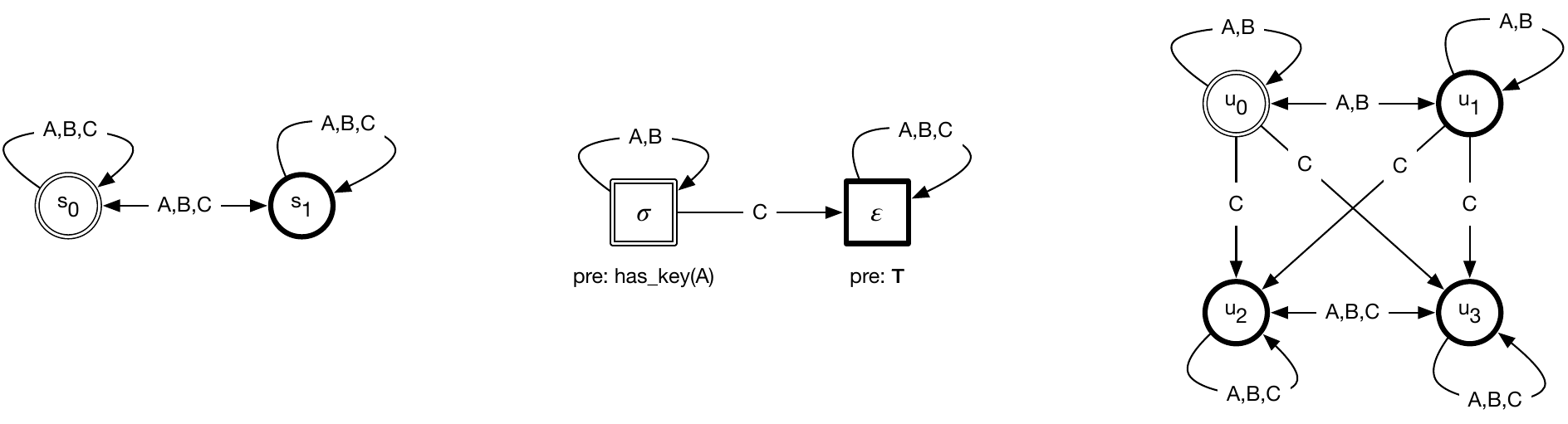}}
\caption{Update instance $(\omega(\textit{open}\langle A \rangle,(\{A,B\},\emptyset,\{C\})),\{\sigma\})$ and its application}
\label{umopen}
\end{figure}

The figure on the right in Figure~\ref{umopen} shows $(M',u_0)$,
the result of the application of the update instance 
to the state $(M,s_0)$ where  \\
\quad $u_0 = (s_0,\sigma)$\footnote{This is to say that $u_0$ denotes the world $(s_0,\sigma)$ as in Definition~\ref{def:updo}.}  
with $M'[\pi](u_0)=\{looking(A),looking(B), has\_key(A),opened\}$ \\
\quad $u_1 = (s_1,\sigma)$ with $M'[\pi](u_1)=\{looking(A),looking(B),head, has\_key(A),opened\}$ \\ 
\quad $u_2 = (s_0,\epsilon)$ with $M'[\pi](u_2)=\{looking(A),looking(B), has\_key(A)\}$ \\
\quad $u_3 = (s_1,\epsilon)$ with $M'[\pi](u_3)=\{looking(A),looking(B),  head, has\_key(A)\}$. 

}
\end{example}
In the next definition, we provide the update instance for 
a sensing or announcement action occurrence  given a frame of reference. For simplicity of
 presentation, we will assume that the set of sensed formulae 
of the action is a singleton.  

\begin{definition}[Update Model/Instance for Sensing/Announcement Actions]
\label{upd-sa}
\label{upd-aa}
Let ${\sf a}$ be a sensing action instance that senses $\varphi$ 
or an announcement action instance that announces $\varphi$ 
with the 
precondition $\psi$ and $\rho=(F,P,O)$ 
be a frame of reference. The update model for ${\sf a}$ and $\rho$,  
 $\omega({\sf a},\rho)$, is defined by $ \langle \Sigma, R_1, \dots, R_n, pre,sub\rangle$  where:
\begin{itemize}
\item[$\circ$] $\Sigma = \{\sigma,\tau,\epsilon\}$; 
\item[$\circ$] $R_i$ is given by  
\[
R_i = \left\{ \begin{array}{lcl}
	\{(\sigma,\sigma), (\tau,\tau), (\epsilon,\epsilon)\} & \hspace{.2cm} & 
				\textit{if } i \in F\\
	\{(\sigma,\sigma), (\tau,\tau), (\epsilon,\epsilon),(\sigma,\tau),(\tau,\sigma)\} &&
	 			\textit{if }i\in P\\
	\{(\sigma,\epsilon),(\tau,\epsilon),(\epsilon,\epsilon)\} && 
				\textit{if } i \in O
	\end{array}\right.
\]
\item[$\circ$] The preconditions $pre$ are defined by
\[
pre(x) = \left\{ \begin{array}{lcl}
	\psi \wedge \varphi	& \hspace{.2cm} & \textit{if } x=\sigma\\
	\psi \wedge \neg \varphi && \textit{if } x = \tau\\
	\top && \textit{if } x = \epsilon
	\end{array}\right.
\]
\item[$\circ$] $sub(x) =\emptyset$ for each $x\in \Sigma$.
\end{itemize}
The update instance for the sensing action occurrence ${\sf a}$ and the frame of reference $\rho$ is $(\omega({\sf a},\rho),\{\sigma,\tau\})$
while 
the update instance for the announcement action occurrence ${\sf a}$ and the frame of reference $\rho$ is $(\omega({\sf a},\rho),\{\sigma\})$
\end{definition}
Observe that an update model of a sensing or announcement action occurrence has three events. 
As we can see, an update model for an announcement action 
and a frame of reference is structure-wise identical to the 
update model for a sensing action and a frame of reference. 
The main distinction lies in the set of designated events in 
the update instance for each type of actions. 
There is only one single designated event for announcement
actions while there are two for sensing actions.  


\begin{example}
{ 
Let us consider the occurrence of $peek\langle A \rangle $ in the state described in Figure~\ref{sensmod} (left).  The 
frame of reference for this occurrence of $peek\langle A \rangle$ is $(\{A\},\{B\},\{C\})$.  
The corresponding update instance is given in Figure \ref{sensmod} (middle). 

\begin{figure}[htbp]
%
\centerline{\includegraphics[width=\textwidth]{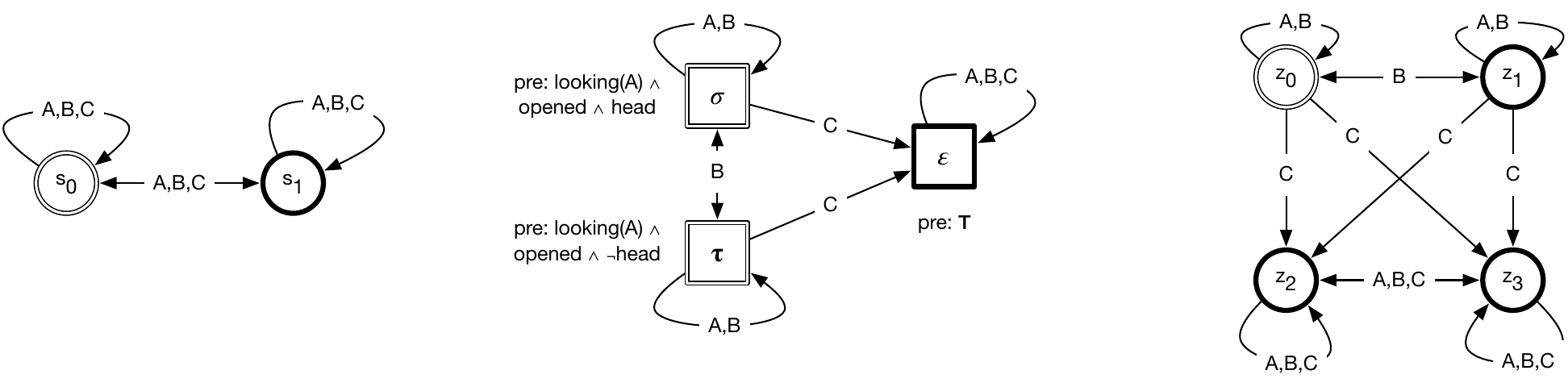}}
\caption{Update instance $(\omega(peek\langle A \rangle, (\{A\},\{B\},\{C\})), \{\sigma,\tau\})$ and its application}
\label{sensmod}
\end{figure}
In the above figure, $z_0=(s_0,\sigma)$, $z_1=(s_1,\tau)$, $z_2=(s_0,\epsilon)$, and $z_3=(s_1,\epsilon)$  
and the interpretation of each world $z_i$ is the same as the interpretation of the world $s_j$ where 
$z_i = (s_j, x)$ for $x \in \{\sigma, \tau, \epsilon\}$.
}
\end{example}

\noindent 
The next example illustrates the update instance of announcement actions.



%

\begin{example} 
{
Let us assume that $A$ and $B$ have agreed to a scheme of 
informing each other if the coin lies heads up by raising 
a hand. $B$ can only observe $A$ if $B$ is looking at the box (or 
looking at $A$). 
$C$ is completely ignorant about the meaning of $A$'s raising her hand.
This can be modeled by the following statements:\footnote{
For simplicity, we ignore the effect that $A$'s hand is raised when $A$ raises her hand.  
}
\[\begin{array}{l}
   \executable  {raising\_hand}\langle A \rangle \iif \B_A(head), head   \\
	{raising\_hand}\langle A \rangle \announces head\\
	A \observes {raising\_hand}\langle A \rangle \iif  \top \\
	B \observes {raising\_hand}\langle A \rangle  \iif looking(B) 
\end{array}
\]
If $A$ knows the coin lies heads up and raises her hand, 
$B$ will be aware  that the coin lies heads up and $C$ is completely 
ignorant about this. 

%
%
%
Let us consider  the action occurrence {\it raising\_hand}$\langle A \rangle$  
and the state in which $B$ is looking 
at the box and thus both $A$ and $B$ are aware of it. 
We have that the  frame 
of reference is $(\{A,B\},\emptyset,\{C\})$ and thus 
the update instance for the occurrence of {\it raising\_hand}$\langle A \rangle$ 
is shown in Figure \ref{ann1mod}.

%

\begin{figure}[htpb]

\centerline{\includegraphics[width=.4\textwidth]{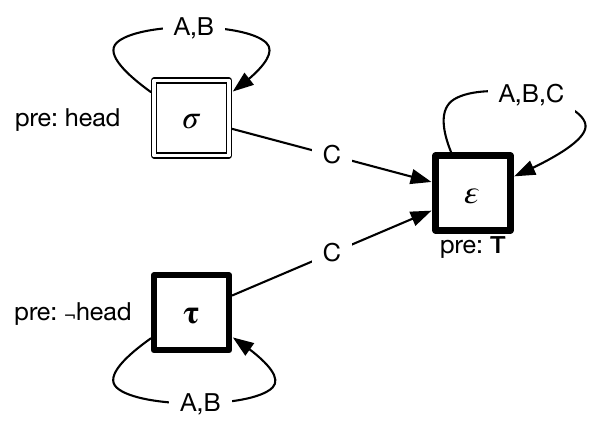}}

\caption{Update instance for the {\it raising\_hand}$\langle A \rangle$ action and $\rho=(\{A,B\},\emptyset,\{C\})$}
\label{ann1mod}
\end{figure}
}
\end{example}

\subsection{Defining $\Phi_D$}

The update models representing action occurrences can be used in formalizing
$\Phi_D$ similar to the proposals in \cite{BaralGPS12,BaralGPS13}. \emph{However,
a main issue of the earlier definitions is that the early definition does not deal with false/incorrect beliefs}. 
Let us address this with a novel solution, inspired by the suggestion in \citep{vEijck17}. Let us 
start by introducing  some
additional notation. For a pointed Kripke model $(M,s)$, an agent $i \in \agents$, and a formula
$\varphi$, we say that $i$ has false belief about $\varphi$ in $(M,s)$ if
$$
(M,s) \models \varphi  \textnormal{ and }  (M,s) \models \B_i \neg \varphi.
$$
For a set of agents $S$,   a pointed Kripke model $(M,s)$, and a formula $\varphi$, such that $(M,s) \models \varphi$,
let $M[S, \varphi]$ be obtained from $M$ by replacing $M[i]$ with $M[S,\varphi][i]$ where

\begin{itemize} 
\item $M[S,\varphi][i] = (M[i] \setminus M[i]^{s}) \cup \{(s,s)\}$ for $i \in S$ and $(M,s) \models \B_i \neg \varphi$ where 
$M[i]^{s} = \{(s,u) \mid (s,u) \in M[i]\}$; and
\item $M[S,\varphi][i] = M[i]$ for other agents, i.e., $i \in \agents \setminus S$ or
  $i \in S$ and $(M,s) \not\models \B_i \neg \varphi$.
\end{itemize}

This process aims at correcting beliefs of agents with false beliefs. 
Intuitively, the links in $M[i]^{s}$ create the false belief of agent $i$. Therefore,  
to correct the false believe of $i$, we should replace them with the single link $(s,s)$. 
Let us now define $\Phi_D$.

\begin{definition} 
Let $D$ be a \mastar{} domain and ${\sf a}$ be an action instance. 
Let $\psi$ be precondition of ${\sf a}$,
$(M,s)$ a state, and $\alpha$ a set of agents.  Let us consider ${\sf a}\in \calai$. We say ${\sf a}$ 
is \emph{executable} in $(M,s)$ if  $(M,s) \models \psi$. The result of executing ${\sf a}$ in
$(M,s)$ is a set of states, denoted by $\Phi_D({\sf a}, (M,s))$  and   defined as follows.
\begin{itemize} 
\item If ${\sf a}$ is not executable in  $(M,s)$   then
$\Phi_D({\sf a}, (M,s)) = \emptyset$

\item If ${\sf a}$ is executable in  $(M,s)$
and $(\cale, E_d)$ is the representation of the occurrence of ${\sf a}$ in $(M,s)$
then
\begin{itemize} 
\item $\Phi_D({\sf a}, (M,s)) =  (M,s) {\otimes} (\cale, E_d)$  if ${\sf a}$ is an world-altering action instance;

\item $\Phi_D({\sf a}, (M,s)) =  M[F_D({\sf a}, M,s), \varphi]  \otimes  (\cale, E_d)$
if ${\sf a}$ is a sensing action instance that senses $\varphi$ and $(M,s) \models \varphi$;

\item $\Phi_D({\sf a}, (M,s)) =  M[F_D({\sf a}, M,s), \neg f]  \otimes  (\cale, E_d)$
if ${\sf a}$ is a sensing action instance that senses $\varphi$ and $(M,s) \models \neg \varphi$;

\item $\Phi_D({\sf a}, (M,s)) =  M[F_D({\sf a}, M,s), \varphi] \otimes  (\cale, E_d)$
if ${\sf a}$ is an announcement action instance that announces $\varphi$.
\end{itemize}
\end{itemize}
Finally, for a set of states $\calm$,
\begin{itemize} 
\item if ${\sf a}$ is not executable in  some $(M,s) \in \calm$ then
$\Phi_D({\sf a}, \calm) = \emptyset$;

\item if ${\sf a}$ is executable in     every $(M,s) \in \calm$ then
$$\Phi_D({\sf a}, \calm) =  \bigcup_{(M,s) \in \calm} \Phi_D({\sf a}, (M,s)).$$
\end{itemize}
\end{definition} 

%
%
%
%

\subsection{Properties of $\Phi_D$} 

While the  syntax and semantics of \mastar{} represent contributions on
their own,  of
particular interest is the fact that \mastar{} satisfies certain useful properties---specifically
its ability to correctly capture certain intuitions concerning the effects of various types of
actions. In particular,
\begin{list}{$\bullet$}{\topsep=1pt \parsep=0pt \itemsep=1pt} 
	\item If an agent is fully observant of the execution of an action instance then 
her beliefs will be updated with the effects of such action occurrence;  
	\item An agent who is only partially observant of the action occurrence
	will believe that the agents who are fully observant of the action occurrence are certain about 
	the action's effects; and 
	\item An agent who is oblivious of the action occurrence will also be ignorant about
	its effects.  
\end{list}
We will next present several theorems discussing these properties. 
To simplify the presentation, we will use the following notations throughout the 
theorems in this subsection.
   
\begin{list}{$\bullet$}{\topsep=1pt \parsep=0pt \itemsep=1pt} 
\item $D$ denotes a consistent \mastar{} domain;
\item $(M,s)$ denotes a state;  and 
\item ${\sf a}$ is an action instance, 
whose precondition is given by the statement $$\executable {\sf a} \iif \psi$$ in $D$, 
and ${\sf a}$  is executable in $(M,s)$. 
\item $\rho = (F,P,O)$ is the frame of reference of the execution of ${\sf a}$ in $(M,s)$ where  
$F = F_D({\sf a}, M,s)$, 
$P = P_D({\sf a}, M,s)$, 
and  $O = O_D({\sf a}, M,s)$.
\end{list} 


We begin with a theorem about the occurrence of the instance of an world-altering action. 

\begin{theorem}
\label{ontic:theorem:1}
Assume that ${\sf a}$ is an world-altering action instance. 
It holds that:
\begin{enumerate}
    \item for every agent $x \in F_D({\sf a}, M,s)$ and 
    $[{\sf a} \causes \ell \iif \varphi]$ belongs to $D$,  
    if $(M,s) \models \B_x \varphi$ then $\Phi_D({\sf a}, (M,s)) \models \B_x \ell$; 

    \item for every agent $y \in O_D({\sf a}, M,s)$ and a belief formula $\eta$, 
    $\Phi_D({\sf a}, (M,s)) \models \B_y \eta$ iff $(M,s) \models \B_y \eta$; and 
   
   \item for every pair of agents $x \in F_D({\sf a}, M,s)$ and $y \in O_D({\sf a}, M,s)$ and a belief formula $\eta$, 
   if $(M,s) \models \B_x \B_y \eta$ then $\Phi_D({\sf a}, (M,s)) \models \B_x \B_y \eta$.  
 
\end{enumerate}
\end{theorem}
\begin{proof} See Appendix A. 
\end{proof}

In the above theorem, the first property discusses the changes in the beliefs of agents who are fully observant of
the occurrence of an world-altering action instance. The second property shows that oblivious agents are still in 
the ``old state,'' i.e., they believe nothing has happened. The third property indicates that fully observant agents 
are also aware that the beliefs of all oblivious agents have not changed. This is particular useful in situations where an 
agent would like to create false beliefs about a fluent $p$ for other agents: she only needs to secretly execute an action
that changes the truth value of $p$.

\begin{theorem}
\label{sensing:theorem:1}
Let us assume that ${\sf a}$ is a sensing action instance 
and  $D$ contains the statement  ${\sf a} \determines f$.
It holds that:
\begin{enumerate}

    \item if $(M,s) \models f$ then $\Phi_D({\sf a}, (M,s)) \models \mathbf{C}_{F_D({\sf a}, M,s)} f$; 
    \item  if $(M,s) \models \neg f$ then $\Phi_D({\sf a}, (M,s)) \models \mathbf{C}_{F_D({\sf a}, M,s)} \neg f$; 

    \item $\Phi_D({\sf a}, (M,s)) \models \mathbf{C}_{P_D({\sf a}, M,s)} (\mathbf{C}_{F_D({\sf a}, M,s)} f \vee 
    								\mathbf{C}_{F_D({\sf a}, M,s)} \neg f)$;   

    \item $\Phi_D({\sf a}, (M,s)) \models \mathbf{C}_{F_D({\sf a}, M,s)} (\mathbf{C}_{P_D({\sf a}, M,s)} (\mathbf{C}_{F_D({\sf a}, M,s)} f \vee 
    								\mathbf{C}_{F_D({\sf a}, M,s)} \neg f))$;   

    \item  for every agent $y \in O_D({\sf a}, M,s)$ and formula $\eta$,  $\Phi_D({\sf a}, (M,s)) \models \B_y \eta$ iff $(M,s) \models \B_y \eta$; 
   
   \item   for every pair of agents $x \in F_D({\sf a}, M,s)$ and $y \in O_D({\sf a}, M,s)$ and a  formula $\eta$ 
    if $(M,s) \models \B_x \B_y \eta$  then $\Phi_D({\sf a}, (M,s)) \models \B_x \B_y \eta$.
  
%
%
%
%
%
%
%

%
%
         
\end{enumerate}
\end{theorem}
\begin{proof} See Appendix A. 
\end{proof}

The first and second properties of the above theorem indicate that agents who are fully aware of
the occurrence of the sensing action instance will be able to update their beliefs with the truth value in the real state of the world of the 
sensed fluent, thereby correcting any false beliefs that they might have before the execution of 
the action. The third property shows that agents who are partially aware of the action execution will 
know that agents who are fully aware of the action execution will have the correct beliefs about the sensed fluents. 
The fourth property indicates that fully aware agents know that partially observed agents would know that they have the correct beliefs about the sensed fluent. The fifth and sixth properties are about oblivious agents' beliefs.  

\begin{theorem}
\label{announcement:theorem:1}
Assume that ${\sf a}$ is an announcement action instance 
and  $D$ contains the statement 
    ${\sf a} \announces \varphi$. 
If $(M,s) \models \varphi $ then it holds that 
\begin{enumerate}
     \item $\Phi_D({\sf a}, (M,s)) \models \mathbf{C}_{F_D({\sf a}, M,s)} \varphi$; 

    \item $\Phi_D({\sf a}, (M,s)) \models \mathbf{C}_{P_D({\sf a}, M,s)} (\mathbf{C}_{F_D({\sf a}, M,s)} \varphi \vee 
    								\mathbf{C}_{F_D({\sf a}, M,s)} \neg \varphi)$;  

    \item $\Phi_D({\sf a}, (M,s)) \models \mathbf{C}_{F_D({\sf a}, M,s)} (\mathbf{C}_{P_D({\sf a}, M,s)} (\mathbf{C}_{F_D({\sf a}, M,s)} \varphi \vee 
    								\mathbf{C}_{F_D({\sf a}, M,s)} \neg \varphi))$;   

    \item  for every agent $y \in O_D({\sf a}, M,s)$ and a formula $\eta$,  $\Phi_D({\sf a}, (M,s)) \models \B_y \eta$ iff $(M,s) \models \B_y \eta$;  and
   
   \item  for every pair of agents $x \in F_D({\sf a}, M,s)$ and $y \in O_D({\sf a}, M,s)$ and  a formula $\eta$,
    if $(M,s) \models \B_x \B_y \eta$ then $\Phi_D({\sf a}, (M,s)) \models \B_x \B_y \eta$.

\end{enumerate}
\end{theorem}
\begin{proof} 
The proof of this theorem is similar to the proof of Theorem~\ref{sensing:theorem:1} and is omitted for brevity. 
\end{proof}

Similarly to Theorem~\ref{sensing:theorem:1}, the first property of the above theorem indicates that a truthful announcement could 
help agents who are fully aware of the action instance occurrence  correct their false beliefs. They also know that 
partially aware agents will know that they have the correct beliefs. 
Likewise, partially aware agents will only know that fully aware agents know the truth value of 
the announced formula but they might not have the real value of this formula themselves. Furthermore, as in 
other types of actions, the beliefs of oblivious agents do not change.   

\subsection{Entailment in \mastar{} Action Theories}
\label{bstate} 


We are now ready to define the notion of entailment in
\mastar{} action theories. It will be defined between 
\mastar{} action theories and queries of the following form: 

\begin{equation} \label{query}
\varphi \after \delta 
\end{equation}
where $\varphi$ is a   formula and $\delta$ is 
a \emph{sequence of action instances} ${\sf a_1};\ldots;{\sf a_n}$ ($n \ge 0$); we will refer to
such type of sequences of action instances as \emph{plans}. 
Let us  observe that the entailment can be easily extended to consider  more general forms 
of \emph{conditional plans,} that include conditional statements (e.g., 
{\bf if-then}) or  loops (e.g., {\bf while})---as  discussed 
in \citep{LevesqueRLLS97,SonB01}. We leave these relatively simple
extensions for future work.

The description of an evolution of a system will deal with  \emph{belief state} (Definition~\ref{def:init_state}). 
For a belief state $B$ and an action instance ${\sf a}$, we say that ${\sf a}$ is executable in $B$ 
if $\Phi_D(\textsf{a},(M,s)) \ne \emptyset$ for every state $(M,s)$ in $B$. With a slight 
abuse of notation, 
we define 
\begin{equation} 
\Phi_D({\sf a}, B) = \left\{
\begin{array}{lcl}
	\{\bot\} & & \textnormal{if $\Phi_D({\sf a},(M,s))=\emptyset$ in some state $(M,s)$ in $B$} \\
	&& \textnormal{or $B=\{\bot\}$} \\
	\bigcup_{(M,s)\in B} \Phi_D({\sf a}, (M,s)) & & \textnormal{otherwise} \\
\end{array}
\right. 
\end{equation} 
where $\{\bot\}$ denotes that the execution of ${\sf a}$ in $B$ fails. Note that we
assume that no action instance is executable in $\bot$.

Let $\delta$ be a plan and $B$ be a belief state. The set of belief states 
resulting from the execution 
of $\delta$ in $B$, denoted by $\Phi_D^*(\delta,B)$, is defined as follows:

\begin{itemize}
\item If $\delta$ is the empty plan $[\: ]$ then $\Phi_D^*([\: ],B) = B$; 

\item If $\delta$ is a plan of the form ${\sf a};\delta'$ (with ${\sf a} \in \calai$),
	then $\Phi_D^*({\sf a};\delta',B) =  \Phi_D^*(\delta', \Phi_D({\sf a},B))$. 
\end{itemize}

Intuitively, the execution of $\delta$ in $B$ can go through 
several paths, each path might finish in 
a set of states. It is easy to see that if one of the 
states reached on a path during the execution
of $\delta$ is $\bot$ (the failed state) then the final result 
of the execution of $\delta$ in $B$ is $\{\bot\}$. 
$\Phi_D^*(\delta,B) = \{\bot\}$ indicates that the execution 
of $\delta$ in $B$ fails.   

We are now ready to define the notion of entailment. 

\begin{definition} 
[Entailment]
An action theory $(I, D)$ entails the query $$\varphi \after \delta$$ denoted by 
$(I, D) \models \varphi \after \delta$
if 
\begin{enumerate} 
\item $\Phi_D^*(\delta, I_0) \ne \{\bot\}$ and 
\item $(M,s) \models \varphi$ for each $(M,s) \in \Phi_D^*(\delta, I_0)$
\end{enumerate} 
where $I_0$ is the initial belief state of $(I, D)$.

We say that $(I, D)$ {\bf S5}-entails the query $\varphi \after \delta$, denoted by
$(I, D) \models_{\mathbf{S5}} \varphi \after \delta$, if 
the two conditions (1)--(2) are satisfied with respect to
 $I_0$ being the initial
{\bf S5}-belief state of $(I, D)$. 
\end{definition}

\subsection{Using \mastar{}: An Illustration}

\noindent
The next example illustrates these definitions. 

\begin{example}
\label{ex11}  
{ 
Let $D_1$ be the domain specification given in 
Examples \ref{ex4} and \ref{ex5}
and $I_1$ be the set of initial statements given in Example \ref{ex3}. 
Furthermore, let $\delta_A$ be the sequence of actions: 
$$\delta_A = distract(C)\langle A \rangle; open\langle A \rangle; peek\langle A \rangle.$$ We can show that 
\[
\begin{array}{l} 
(I_1, D_1) \models_{\mathbf{S5}} (\B_A head \vee \B_A \neg head) \wedge \B_A(\B_B (\B_A head \vee \B_A \neg head))  \after \delta_A  \\
(I_1, D_1) \models_{\mathbf{S5}} \B_B (\B_A head \vee \B_A \neg head) \wedge  \neg \B_B head \wedge \neg \B_B \neg head \after \delta_A \\
(I_1, D_1) \models_{\mathbf{S5}} \B_C [\bigwedge_{i \in \{A,B,C\}} (\neg \B_i head \wedge \neg \B_i \neg head)] \after \delta_A 
\end{array} 
\]


It can be shown that $(I_1,D_1)$ is indeed a definite action theory and 
any {\bf S5}-initial state of $(I_1,D_1)$ is equivalent to either $(M_0,s_0)$ or $(M_0,s_1)$ 
where $(M_0,s_0)$ is drawn in Figure~\ref{ex-init-png} (left) 
and 
$M_0[\pi](s_0) = \left\{ has\_key(A), looking(A), looking(B), looking(C) \right\}$ 
and   \\
$M_0[\pi](s_1) = \left\{ has\_key(A), looking(A), looking(B), looking(C), head  \right\}.$

The execution of $distract(C)\langle A \rangle$ in $(M_0,s_0)$ 
results in a new state $(M_1,u_0)$ and is shown in Figure~\ref{ex-init-png} (right). 
The update model corresponds to 
the occurrence of $distract(C)\langle A \rangle$ in $(M_0,s_0)$ is shown in the middle 
of  Figure~\ref{ex-init-png},  and  the interpretations associated with the worlds in $M_1$ are: \\
\quad $u_0 = (s_0,\sigma)$
with $M_1[\pi](u_0)=\{looking(A),looking(B), has\_key(A)\}$ \\
\quad $u_1 = (s_1,\sigma)$ with $M_1[\pi](u_1)=\{looking(A),looking(B),head,has\_key(A)\}$ \\ 
\quad $u_2 = (s_0,\epsilon)$ with $M_1[\pi](u_2)=\{looking(A),looking(B),looking(C), has\_key(A)\}$ \\
\quad $u_3 = (s_1,\epsilon)$ with $M_1[\pi](u_3)=\{looking(A),looking(B),looking(C),head,has\_key(A)\}$ 

\begin{figure}[htbp]
\centerline{\includegraphics[width=.80\textwidth]{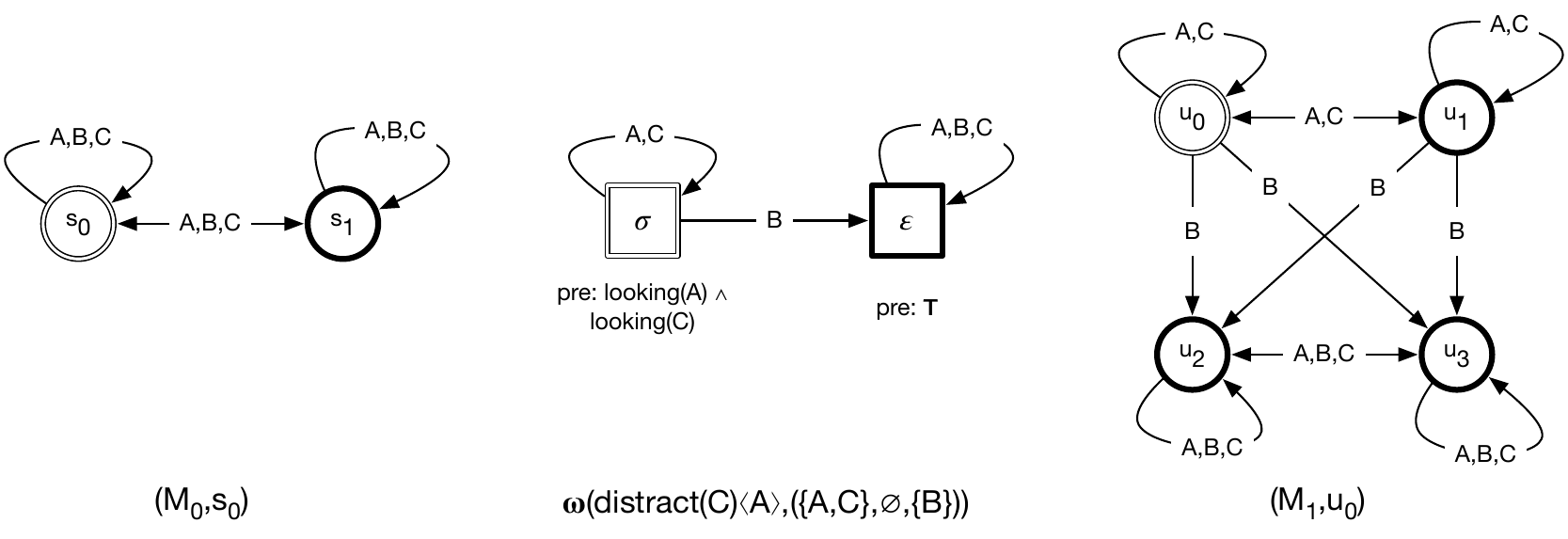}}
\caption{Execution of $distract(C)\langle A \rangle$ in $(M_0,s_0)$ results in $(M_1,u_0)$}
\label{ex-init-png} 
\end{figure}
  
The execution of $open\langle A \rangle$ in $(M_1,u_0)$ (left, Figure~\ref{ex-init-open}) results in a new state $(M_2,v_0)$ 
(right, Figure~\ref{ex-init-open}). The update model corresponds to 
the occurrence of $open\langle A \rangle$ in $(M_1,u_0)$ is shown in the middle of Figure~\ref{ex-init-open}. The interpretations 
associated to each world of  $(M_2,v_0)$ are as follows: \\
\quad $v_0 = (u_0,\sigma)$ with $M_2[\pi](v_0)=\{looking(A),looking(B), has\_key(A),opened\}$ \\
\quad $v_1 = (u_1,\sigma)$ with $M_2[\pi](v_1)=\{looking(A),looking(B),head,has\_key(A),opened\}$ \\ 
\quad $v_2 = (u_2,\sigma)$ with $M_2[\pi](v_2)=\{looking(A),looking(B),looking(C), has\_key(A),opened\}$ \\
\quad $v_3 = (u_3,\sigma)$ with $M_2[\pi](v_3)=\{looking(A),looking(B),looking(C),head,has\_key(A),opened\}$ \\
\quad $v_4 = (u_0,\epsilon)$ with $M_2[\pi](v_4)=\{looking(A),looking(B), has\_key(A)\}$ \\
\quad $v_5 = (u_1,\epsilon)$ with $M_2[\pi](v_5)=\{looking(A),looking(B),head,has\_key(A)\}$ \\ 
\quad $v_6 = (u_2,\epsilon)$ with $M_2[\pi](v_6)=\{looking(A),looking(B),looking(C), has\_key(A)\}$ \\
\quad $v_7 = (u_3,\epsilon)$ with $M_2[\pi](v_7)=\{looking(A),looking(B),looking(C),head,has\_key(A)\}$ 
\begin{figure}[htbp]
\centerline{\includegraphics[width=\textwidth]{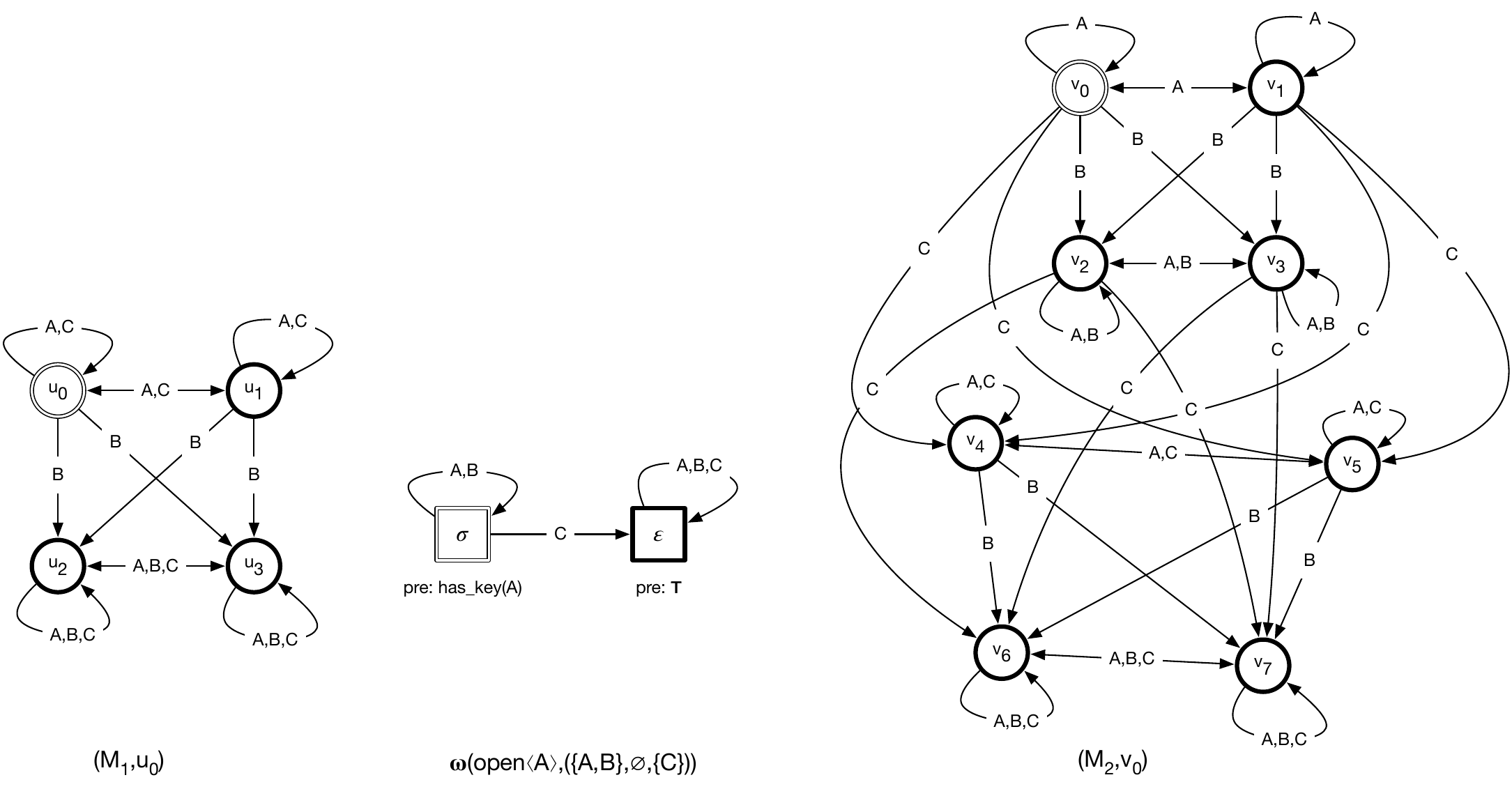}}
\caption{Execution of $open\langle A \rangle$ in $(M_1,u_0)$ results in $(M_2,v_0)$}
\label{ex-init-open} 
\end{figure}

Finally, the execution of $peek\langle A \rangle$ in $(M_2,v_0)$ results in $(M_3,z_0)$ (Figure~\ref{ex-init-peek})
where $z_0 = (v_0,\sigma)$,  $z_1 = (v_2,\sigma)$, $z_2 = (v_1,\tau)$, $z_3 = (v_3,\sigma)$,
and $z_{i+4} = (v_i, \tau)$ for $i=0,\ldots,7$ 
with $M_3[\pi](z_0) = M_2[\pi](v_0)$,  $M_3[\pi](z_1) = M_2[\pi](v_2)$,
 $M_3[\pi](z_2) = M_2[\pi](v_1)$,  $M_3[\pi](z_3) = M_2[\pi](v_3)$, 
 and  $M_3[\pi](z_{i+4}) = M_2[\pi](v_i)$ for $i=0,\ldots,7$.

\begin{figure}[htbp]
\centerline{\includegraphics[width=\textwidth]{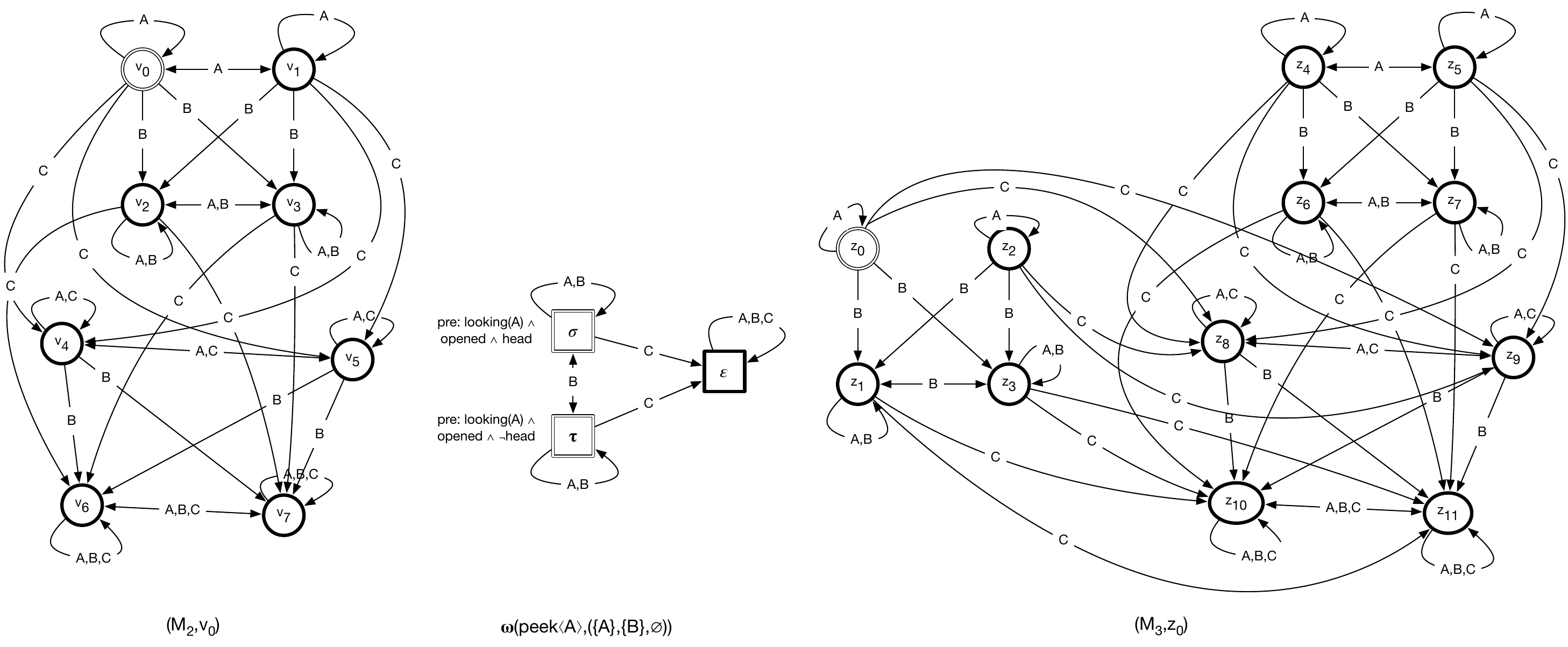}}
\caption{Execution of $peek\langle A \rangle$ in $(M_2,v_0)$ results in $(M_3,z_0)$}
\label{ex-init-peek} 
\end{figure}
It is easy to see that the execution of $\delta_A$ in $(M_0,s_0)$ results in only one state $(M_3,z_0)$. We can verify that 
\begin{equation} \label{ex-m0-s0}
\begin{array}{l} 
(M_3,z_0) \models  \B_A \neg head \wedge \B_A(\B_B (\B_A head \vee \B_A \neg head))    \\
(M_3,z_0) \models  \B_B (\B_A head \vee \B_A \neg head) \wedge  (\neg \B_B head \wedge \neg \B_B \neg head)   \\
(M_3,z_0) \models  \B_C [\bigwedge_{i \in \{A,B,C\}} (\neg \B_i head \wedge \neg \B_i \neg head)]  
\end{array} 
\end{equation}

We conclude the example with a note that  the execution of $\delta_A$ in $(M_0,s_1)$ results in a state with minor differences 
comparing to $(M_3,z_0)$ and is shown in Figure~\ref{ex-init-m0-s1}.   

\begin{figure}[htbp]
\centerline{\includegraphics[width=\textwidth]{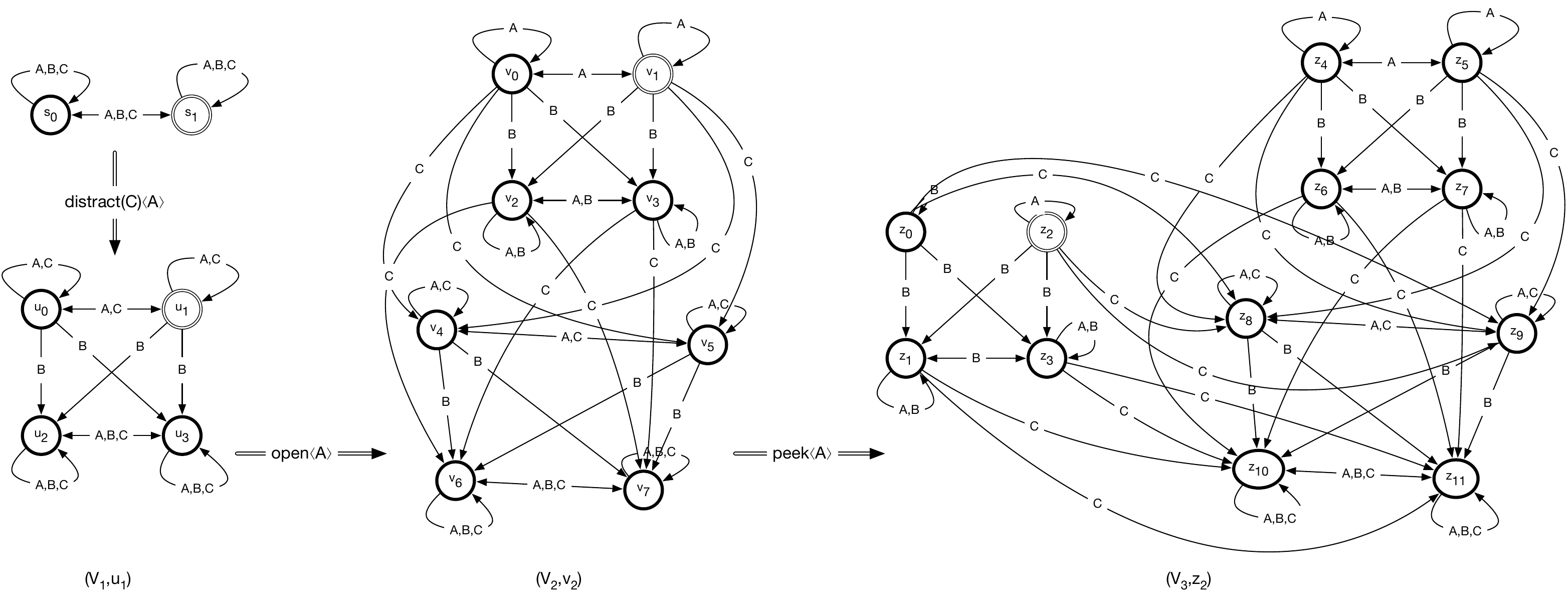}}
\caption{Execution of $\delta_A$ in $(M_0,s_1)$ results in $(V_3,z_2)$}
\label{ex-init-m0-s1} 
\end{figure}
  
We can also verify that the following holds:
\begin{equation} \label{ex-m0-s1}
\begin{array}{l} 
(V_3,z_2) \models  \B_A head \wedge \B_A(\B_B (\B_A head \vee \B_A \neg head))    \\
(V_3,z_2) \models  \B_B (\B_A head \vee \B_A \neg head) \wedge  (\neg \B_B head \wedge \neg \B_B \neg head)   \\
(V_3,z_2) \models  \B_C [\bigwedge_{i \in \{A,B,C\}} (\neg \B_i head \wedge \neg \B_i \neg head)]  
\end{array} 
\end{equation}

\eqref{ex-m0-s0} and \eqref{ex-m0-s1} prove the conclusions of the example. 

}
\end{example}

\section{Related Work and Discussion}
\label{related}

In this section, we connect our work to related efforts in
reasoning about actions and their effects in multi-agent
domains. The background literature  spans multiple
areas. We will give a quick introduction  and
focus our attention on the most closely related works.

\subsection{Relating \mastar{} and DEL- and Update Model-based Languages}

Throughout the paper, we refer to update model and employ it in defining the semantics of \mastar{}. The reasons are twofold. First, the use of update models has been well-accepted in reasoning about dynamic multi-agent systems. Second, formulae in update model language can be translated into DEL formulae without action modalities (Chapter 7, \cite{vanDitmarschHK07}). For this reason, we will simply use ``the update model language'' or DEL to refer to both update model- or DEL-based language.

The key distinction between \mastar{} specification and DEL specification lies in the fact that the specification of effects of actions in \mastar{} emphasizes the distinction between \emph{direct} and \emph{indirect} effects of an action occurrence while DEL specification focuses on \emph{all} effects. To reiterate, in \mastar{},

\begin{itemize}
\item the direct effects of a world-altering action occurrence are the changes of the world that are directly caused by the execution of an action; for instance, the direct effect of $open\langle A \rangle$ is the box is opened;

\item the direct effects of a sensing action occurrence are the truth values of the sensed formula; for instance, the direct effect of $peek\langle A \rangle$, when the coin lies tails up, is $tail$ is true;

\item the direct effect of a truthful announcement action occurrence is that the truth value of the announced formula is true; for instance, the direct effect of $shout\_tail\langle A \rangle$ is that the information ``$tail$ is true'' was announced.
\end{itemize}
In all cases, the indirect effects of an action occurrence are the changes in the beliefs of agents who might be aware of the action occurrence. Such indirect effects are specified by the set of statements of the form \eqref{observes} and \eqref{pobserves}. We note that even in single agent domains world-altering actions could also create indirect effects and this has been well studied (see, e.g., \cite{GiunchigliaKL97,Shanahan99}). The advantage of directly dealing with indirect effects---in the form of state constraints---in planning has been discussed in \cite{ThiebauxHN03,TuSGM11}. This type of indirect effects of actions could be included in \mastar{} by adding statements, referred as \emph{static causal laws}, of the form
\[
\varphi \iif \psi
\]
where $\varphi$ and $\psi$ are fluent formulae. We have conducted a preliminary investigation of this issue in an earlier version of \mastar{} in \cite{BaralGPS13}. We decide to keep it out of this paper for the simplicity of the presentation.

Contrary to the design of \mastar{}, formulae in update model language aim at specifying \emph{all effects} of an action occurrence. To see it, let us consider the action $peek\langle A\rangle$. It is generally accepted that this action helps $A$---under the right circumstance---learn whether or not the coin lies heads or tails up. Since the effects of this action on the knowledge of other agents are different in different situations (pointed Kripke structure), different formulae will need to be developed to describe the effects of this particular action (more on this in Example~\ref{ex:size-update-model-ma}).  In \mastar{}, such indirect effects of action occurrences are encoded in statements of the form \eqref{observes} and \eqref{pobserves}. Due to the restrictions imposed on these two types of statements, some different effects of actions that could be represented using update models cannot be described in \mastar{} (c.f., Sub-subsection~\ref{subsub:expressivity}).

The reader should note that in this discussion \emph{two different kinds of objects} are touched on: \emph{action occurrences} and \emph{actions}. This is intentional -- in the update model language, the \emph{kinds of objects} being discussed are rightly thought of as action occurrences (i.e., individual members of a more abstract type); whereas in \mastar{} (and action language approach more generally), the fundamental abstraction is that of an \emph{action} (seen as a kind of type, of which distinct occurrences are members). The difference in focus leads to a much greater simplicity of action descriptions in presence of multiple agents.
Let us consider the simplified version of the coin in a box problem as presented in
Example \ref{ex2}---with  three agents $A$, $B$, and $C$,  a box containing a coin, and initially it is common knowledge that none
of the agents knows whether the coin lies heads up or tails up. Let us assume that
 agent $A$ peeks into  the box. In our formalism, we express the action of $A$ as \emph{an instance} $peek\langle A \rangle$ of the action \emph{peek}.
 In DEL, the update model \emph{for the same action occurrence}, as given in Figure~\ref{sensmod},
 will also include additional information about all three agents $A$, $B$, and $C$ encoding their ``roles'' or ``perspectives''
  while $A$ is peeking into the box
 (more in Sub-subsection~\ref{subsub:compactness}).
  By roles or perspectives we mean information about what the agents are doing, in terms of who is watching whom and who knows about that.

 It is evident that our representation of the action simply as
 $peek\langle A \rangle$ is much simpler than the DEL representation in Figure~\ref{sensmod}.
 But our representation does not include the information about what else the agents $A$, $B$, and $C$ are doing while $A$ is peeking into the box.
 In our formulation, such information is a \emph{part of the state,} and is expressed by using perspective fluents,  such as
 \emph{looking($B$)}---that encodes the information that $B$ is looking at the box---and
 \emph{group\_member($B$,group($A$))}---that encodes the information that $B$ and $A$ are  together in the same group.

Thus, it appears that a critical difference between the \mastar{} approach to representing multi-agent actions and the approach
used in DEL with update models lies in the way
 we encode the information about agents roles and perspectives---as part of the action in DEL with update models and
 as part of the state in \mastar{}.  There are some important implications of such difference and we discuss them in the following subsections.


\subsubsection{Compactness of Representation}
\label{subsub:compactness}

As we discussed in Section~\ref{sec:updatemodel},  each action occurrence in \mastar{} corresponds to an update model. As such, any \mastar{} domain could be represented as a theory in update model language. Since each action occurrence is associated with a pointed Kripke structure, it follows that, theoretically, we might need \emph{infinitely many formulae} for representing a \mastar{} domain.
The next example shows that we need an exponential number of update models to describe an action with linear number of ``indirect effects'' in  \mastar{} domain.

\begin{example}
\label{ex:size-update-model-ma}
Let us consider, for example, the action occurrence 
{\it peek$\langle A \rangle$} from domain $D_1$. To describe different scenarios related to this action occurrence,  we  need to have an update model for all of
the following cases:
\begin{list}{$\bullet$}{\topsep=1pt \parsep=0pt \itemsep=1pt}
\item  Both $B$ and $C$ are looking;
\item  Either $B$ or $C$ is looking but not both; and
\item  Both $B$ and  $C$ are not looking.
\end{list}
In our approach, the above example is specified in a very different way: the action is about
sensing $head$ (or $tail$). The agents who sense it, who observe the sensing take place, and who are oblivious can be specified
directly or can be specified indirectly in terms of \emph{conditions,}
such as which agents are near the sensing, which ones are watching from far,
 and which ones are looking away, respectively.
 As such, to specify all four cases, we write
 \[
B \pobserves peek\langle A \rangle  \iif looking(B)
\quad \textnormal{ and } \quad
C \pobserves peek\langle A \rangle  \iif looking(C).
 \]

It is easy to see that if we have $n$ agents and agent $A$ executes the action $peek$ then
if we want to include the various possibilities in the finite set of action models that can be used for planning
in the context of \cite{BolanderA11},  we will need $2^{n-1}$ action models 
for specifying all possible consequences of $peek\langle A \rangle$.
On the other hand, we only need $n-1$ statements of the form \eqref{pobserves} in the action specification part,  
as the set of looking(X) in our framework is part of the state.
\end{example}

It is easy to see that similar conclusions can be made with regards to an occurrence of a world-altering action or an announcing action. In summary, we can say that to represent a \mastar{} domain $D$ in the action model language, an exponential number of action models is needed. This advantage of \mastar{} can also be seen in representing and reasoning about action sequences. We observe that extensions of DEL have been proposed to address this issue in~\cite{Bolander14} and \cite{EngesserMNT18}.

%
%
%
%
%


\paragraph{Narratives and Dynamic Evolution of Multi-agent Actions}
Let us  consider a scenario with  two agents $A$ and $B$.
Initially, agent $B$ is looking at agent $A$. Agent $A$
lifts a block and, after some time, agent $A$ puts down the block. Some time later, agent $B$ is
distracted, say by $A$, and then agent $A$ again lifts the block.

In our formulation, this narrative can be formalized by first describing the initial situation,
 and then describing the sequence of actions that occurred, which for this example is:
 $$ \textit{liftBlock} \langle A \rangle;  \textit{putDown}\langle A \rangle;  \textit{distract}(B) \langle A \rangle ; \textit{liftBlock}\langle A \rangle.$$
 The description of this evolution of scenario in DEL is not as simple: each action occurrence will have to be
 described as an update model containing information about both agents $A$ and $B$. In addition, such a description
 (in DEL) will be partly superfluous, as it will have to record information about $B$ looking (or not looking) at $A$
 in the update model, when that information is already part of the state.
 Thus, the approach used in \mastar{} to describe this narrative is more natural than the representation in DEL.

Observe that, in our narrative, the action instance  {\it liftBlock$\langle A \rangle $} appears twice. However, due to the difference
in the roles and perspectives over time, the two occurrences of  {\it liftBlock$\langle A \rangle$} correspond to two different
update models. This shows how, using the \mastar{}  formulation, we can support the dynamic evolution of update
models, as result of changes in perspective fluents in the state.
In DEL, the two update models are distinct and there is no direct connection between them and neither one does  evolve from the other.

In order to further reinforce this point, let  us consider another narrative example.
Let us consider a scenario with three agents, $A$, $B$, and $C$. Initially, it is
 common knowledge that none of the agents knows whether the coin in the box is  lying heads up or tails up.
 In addition, let us assume that initially $A$ and $B$ are looking at the box, while $C$ is looking away.
 Let us consider the narrative where $A$ peeks into the box; afterwards, $A$ realizes that $C$ is distracted and
 signals $C$ to look at the box as well; finally $A$ peeks into the box one more time. In \mastar{}, this situation can be
 described again by a sequence of actions:
 $$ \textit{peek}\langle A \rangle; \textit{signal}(C)\langle A \rangle; \textit{peek}\langle A \rangle$$
 The two occurrences of {\it peek($A)$} correspond to two different update models; the second
 occurrence is an evolution of the first caused by the execution of {\it signal(C)$\langle A \rangle$.}
 In DEL, the relevance of the intermediate action {\it signal($C$)$\langle A \rangle$}, and its impact on the second
 occurrence of {\it peek$\langle A \rangle$},  is mostly lost---and this results in the use of two distinct update models
 for $ peek\langle A \rangle$ with complete information about the whole action scenario.

The key aspect that allows a natural representation of narratives and evolution of update models in
\mastar{} is the presence of  the agents' perspectives and roles encoded as
perspective fluents of a state,  and their use to dynamically generate the update
 models of the actions. While DEL can include perspective fluents as part of the states as well,
 it does not have a way to take advantage of them in a similar way as \mastar{}.  
 

%

\subsubsection{Simplicity vs. Expressivity}
\label{subsub:expressivity}

The formulation adopted in this paper is limited in expressivity to ensure simplicity.
It is limited by the (perspective) fluents we have and how we use them.  On the other hand,
the action model language is more complex and also more expressive. This
is also evident in the simplicity of the update models used in \mastar{} (Definitions~\ref{upd-wa}--\ref{upd-aa}).
This leads to the following limitations of \mastar{} in comparison with the action model language:

\paragraph{Complex action occurrences}
\label{a:imaginative}

The simple version of \mastar{} as presented here does not consider complex epistemic action occurrences. An example of this type of action occurrences is the \textbf{mayread} action in Example 5.4   from~\cite{vanDitmarschHK07}. It represents an event that might or might not happen.  Let us have a closer look at the action \textbf{mayread}. \emph{A}nne and \emph{B}ill are in a cafe. Some agent brings a letter to Anne. The letter says that  United Agents is doing well. \emph{B} leaves the table and orders a drink at the bar so that  \emph{A} may have read the letter while he is away. In this example,  \emph{A} did not read the letter.

\begin{figure}[htbp]
\centerline{
     \includegraphics[width=.65\textwidth]{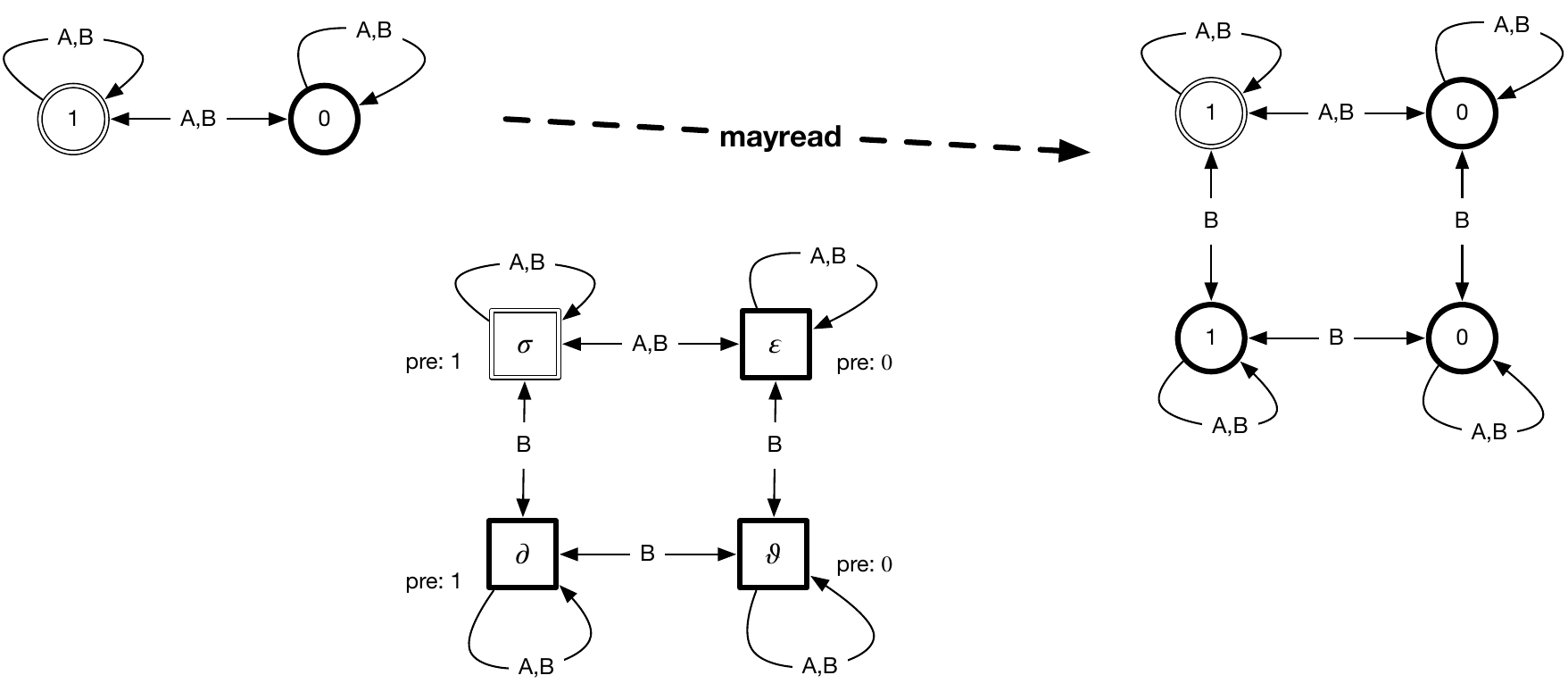}
}
\caption{Update model of \textbf{mayread} without an equivalent \mastar{} representation}
\label{a:mayread}
\end{figure}
Initially, both \emph{A} and \emph{B} do not know the content of the letter sent to \emph{A}.
Figure~\ref{a:mayread} depicts the initial pointed Kripke structure (left) and the pointed Kripke structure after \textbf{mayread} (right).
An update model representing the action \textbf{mayread} is given in the middle of Figure~\ref{a:mayread}.
It is easy to see that any update model facilitating this transition will need to have at least four events, two with 1 as the precondition and two with 0 the precondition. Since any update model in \mastar{} needs at most three events, this shows that there exists no equivalent \mastar{} representation of \textbf{mayread}.

In our action language based framework, \textbf{mayread} can be viewed as a combination (disjunction) of two ``primitive actions,'' one is \emph{A reads the letter} and another one is \emph{A does nothing}. One can then use  constructs from Golog \cite{LevesqueRLLS97} with \mastar{} actions as primitive actions to express  \textbf{mayread}. 

We note that \mastar{} also does not consider non-deterministic world-altering actions  (e.g., the action of tossing a coin results in the coin lies heads or tails up). This type of actions has been extensively studied in action languages for single-agent domains (see, e.g., \cite{GelfondL98}). They can be added to \mastar{} easily by allowing $\ell$ in \eqref{causes} to be arbitrary formula. Similar to static causal laws, we decide to keep it out of this paper for the simplicity of the presentation in this paper.

\begin{figure}[htbp]
\centerline{
     \includegraphics[width=.75\textwidth]{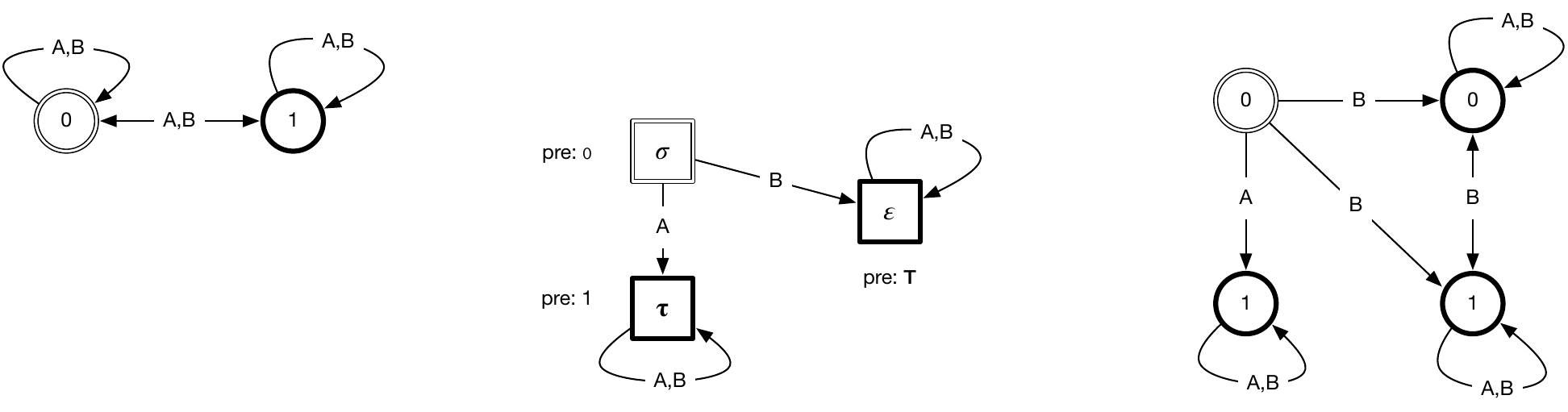}
}
\caption{An update model of an action requiring different type of observability statements without an equivalent \mastar{} representation}
\label{model-belief}
\end{figure}

\paragraph{Agents' Observability}
In \mastar{}, statements of the form \eqref{observes} and \eqref{pobserves} are used for specifying an agent's observability of action occurrences. It is expressed via fluent formulae and is evaluated with respect to the real-state of the world. In general, such observability can be beliefs of the agent about other agents. Consider the update model\footnote{
    We thank an anonymous reviewer of an earlier version of this paper who
    suggested a similar example.
} in Figure~\ref{model-belief} (middle).
It represents an action $\alpha$ of an agent $A$ who believes that after she executes the action then both $A$ and $B$
can see an incorrect outcome $1$---i.e., $A$ and $B$ are fully observant. In reality, $B$ is oblivious.
The initial pointed Kripke structure is given in the left and the result of $\alpha$ is on the right of Figure~\ref{model-belief}.
This shows that, in multi-agent domains, an agent's observability
could also be considered as beliefs, and as such affect
 the beliefs of an agent about other agents'
beliefs after the execution of an action.  The present \mastar{} language does not allow
for such specification.

The simplicity of our formulation is by design, and not an inherent flaw of our approach.  Indeed,
one could envision developing a complete encoding of the complex graph structure of an update model
as part of state, using an extended collection of perspective fluents---but, at this time,
we do not have a corresponding theory of change to guide us in using these more
expressive perspective fluents to capture the full expressive power of update models in DEL. Hence,
our current formalism is less expressive than DEL.
However, the higher expressiveness of update models provides us with a target to expand
\mastar{} and capture more  general actions.

\subsubsection{Other Differences}

The previous sub-subsections detail the key differences between \mastar{} and DEL based on the differences in design and focus of \mastar{} and DEL. We  next discuss their differences from other angles.

\paragraph{Analogy with Belief Update}
Another approach to explore the differences between \mastar{} and DEL builds on the
analogy to the corresponding differences between \emph{belief updates}
and the treatment of actions and change in early action languages
 \citep{GelfondL98}.

Papers on belief updates define and study the problem of updating a formula $\phi$ with a formula $\psi$.
In contrast, in reasoning about actions and change, the focus is on defining the resulting state of the world after
a particular action is performed in a particular world, given a description of {\em (i)} how the action may change the world,
{\em (ii)} when the action can be executed; and
{\em (iii)}  how the fluents in the world may be (possibly causally) related to each other.
In such a context, given a state $s$ and an action $a$, it is possible to see the determination of the resulting state
as the update of $s$ by a formula $\varphi$; But, what is important to consider is that the $\varphi$ is not just the collection
of effects of the action $a$, but incorporates several other components, that take into account the static causal laws
as well as which conditions (part of the conditional effects of $a$) are true in $s$.

This situation is not dissimilar to the distinction between DEL update models and \mastar{}. An update model can be encoded
by an action formula, and the resulting state can be obtained by updating the starting state with such formula. In DEL, such
action formula has to be given directly. Instead,  our considerations in \mastar{} are in the spirit of the early research in reasoning about actions
and change---where we focus on describing actions and their effects, their executability conditions, and where a resulting
``state'' is determined by applying these descriptions to the ``state'' where a particular action is performed. Thus, the action formula
in this latter case is not explicitly given, but derived from the description of the actions, their effects, and executability conditions.

Taking the analogy further, an important application of reasoning about actions is to determine action sequences
or plan structures that achieve a given goal.  This is different from searching for a sequence of formulae $\psi_i$'s which, if used to sequentially update a given initial state, will generate  a goal state.



\paragraph{Executing Actions}
The  notion of actions adopted in \mastar{} is designed to enable their executions  by one or multiple agents. For example,
the action instance \emph{peek$\langle A \rangle$} can be executed only by agent $A$.
On the other hand, the notion of an update model is designed for describing the state changes 
and does not include the information about the actors of the update model, i.e., the agents who 
will execute the actions specified by the model. 
It is therefore not always possible to identify the agents who would execute an update model from its description. 
For example, by simply looking at Figure~\ref{ann1mod} or examining the definition of the corresponding update model, 
we cannot distinguish whether the update model is about the instance $raising\_hand(A)$ or $raising\_hand(B)$. 
In the planning using DEL setting, relation between agents and action models is introduced in \cite{EngesserBMN17,LoewePW11} 
to address this issue.

 

Now consider perspective fluents: how does one execute the perspective fluents, such as \emph{looking($B$)}? The answer is that they are fluents, and they are not required or supposed to be executed.
A more appropriate question would be: how do they become true? The answer is that, while our formulation could have
some actions that make them true, when describing a state we do not need to worry about how exactly the facts in the state
came about to be true. This is not the case when describing actions: when describing actions we need to describe
something that can be executed. In summary, actions, or parts of actions, are supposed to be something that can be executed,
while states, or parts  of states, do not have such requirement.

Hence, our representation of actions where perspective fluents are part of the state (and not part of the action) is more appropriate, and follows the common meaning of an action,\footnote{For example, the relevant dictionary meaning of ``action is (1) something done or performed; act; deed. (2) an act that one consciously wills and that may be characterized by physical or mental activity.}
than the representation of actions in the initial formulations of DEL \cite{vanDitmarschHK07}. 

\paragraph{Value of Update Models:}
Having discussed the differences between \mastar{} and update models, we would like to point out
that update models present a very good technical tool for the understanding of
effects of actions in multi-agent domains. In fact, the transition function $\Phi$ for
\mastar{} action theories can be effectively characterized using update models, as
described in Section~\ref{sec:updatemodel}.

%
%
%

\subsection{Previous Work by the Authors}

Early attempts to adapt action languages to  formalize multi-agent
domains can be found in \citep{BaralSP09a,SonPS09,SonS09b}. In these works, the
action languages $\cal A$, $\cal B$, and $\cal C$ have been extended to formalize
multi-agent domains.

The works in \citep{SonPS09,SonS09b} investigate the use
of action language in multi-agent planning context and focus on the generation
of decentralized plans for multiple agents, to either jointly achieve a goal or
individual goals.

In \citep{BaralSP09a}, we show that several examples found
in the literature---created to address certain aspect in multi-agent systems
(e.g., \citep{BoellaT05,Gerbrandy06,HoekJW05,HerzigT06,SauroGHW06,SpaanGV06})---can
be formalized using an extension of the action language $\cal C$.
Yet, most of the extensions considered in \citep{BaralSP09a,SonPS09,SonS09b} are inadequate for
formalizing multi-agent domains in which reasoning about knowledge of other
agents is critical. To address this shortcoming, we developed and investigated
several preliminary versions of \mastar{} \citep{BaralGPS10a,BaralGPS10b,BaralG10}. We started with
an attempt to formulate knowledge of multiple agents in \citep{BaralGPS10a}; we
successively extended this preliminary version of \mastar{} with the use
of static
 observability specifications in \citep{BaralGPS10b}. The language developed
in this paper subsumes that of \citep{BaralGPS10b}.
In \citep{BaralG10}, we demonstrated
the use of update models to describe the
transition function for the action language of \citep{BaralGPS10b}.

\mastar{} differs from all earlier versions of the language in that its clearly
differentiates the agents who execute the action from the agents who would
observe (or partially observe) the effects of the actions, or are oblivious of
the action execution. This is important since an agent might execute an
action without observing the effects of the action. For example, a blind
agent would not be able to observe the effects of his action of switching
the contact; a agent firing the gun in a pitch dark night would not be able to
observe whether or not he hits the target.

\begin{figure}[htbp]
\centerline{
     \includegraphics[width=\textwidth]{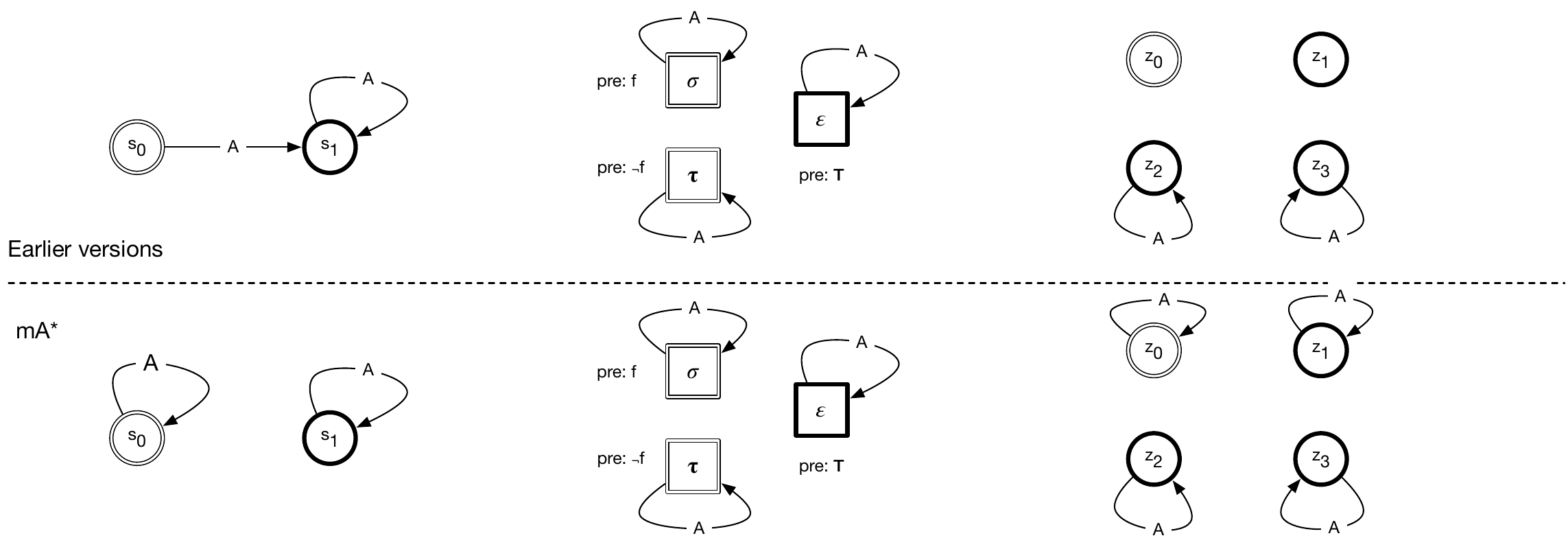}
}
\caption{Sensing $f$ helps $A$ to correct her false beliefs in \mastar{}}
\label{correcting-belief}
\end{figure}

Another, much more important,
difference between \mastar{} and earlier versions of the language lies in
the definition of the function $\Phi_D$. In earlier versions, sensing (or announcement)
actions \emph{do not} help the agents in correcting their beliefs. This can be seen in Figure~\ref{correcting-belief}.
In this example $A$ has the false belief about $f$ ($f$ is true in $s_0$, the real state of the world, and false in $s_1$).
$A$ executes the action that senses $f$. The top part shows how earlier versions of \mastar{} treat this sensing action
occurrence whose update instance is in the middle of the figure. The result is the state shown on the right with four
disconnected worlds $z_0 = (s_0,\sigma)$, $z_1 = (s_0,\epsilon)$, $z_2 = (s_1,\tau)$, and $z_3 = (s_1,\epsilon)$
with the interpretation of $z_j$ identical to that of $s_i$ where $z_j = (s_i,\_)$. As we can see,
$A$ becomes ignorant about everything in this state.
The bottom part shows how \mastar{} deals with such situation: first, it corrects the beliefs of $A$ and then applies the
update. This results in the state on the right (bottom) in which $A$ knows that $f$ is true, which corresponds to the
intuitive result of sensing.

\subsection{Other Languages}

A generalized version of STRIPS, called MA-STRIPS,  has been proposed for
studying multi-agent classical planning model \citep{BrafmanD08}. In this model, each agent possess a set of actions that
they can execute. Distinctions between public and private fluents are considered,
which allow for the definition of internal and public actions. This extension is closely related to
our earlier extensions of action languages (e.g., \citep{BaralSP09a,SonPS09}). Therefore,
the key difference between \mastar{} and MA-STRIPS lies in the focus on reasoning about beliefs of agents
about the world and about the beliefs of other agents in \mastar{} that is not considered in MA-STRIPS. On
the other hand, the focus in \citep{BrafmanD08} is to develop plans for multiple agents to coordinate in achieving
a give state of the world, which is not our focus in the development of \mastar{}.

%
%

GDL-III, introduced in \citep{Thielscher17}, as an epistemic game specification language,  
could potentially be used as a specification language for epistemic planning.
Syntactically,  GLD-III includes specification for two predicates for
reasoning about knowledge of each agent \emph{knows(r,p)}
and common knowledge \emph{knows(p)} among all agents as well as sensing actions (e.g., the action \emph{peek} in Example~\ref{ex1})
and announcement actions (e.g., the action \emph{shout\_tail}  in Example~\ref{ex1}). The effects of actions include \emph{observations}
that allow for the agents to update their knowledge.
The semantics of a game specification in GDL-III is defined
over knowledge states and sequences of actions,
each knowledge state is a pair of a set of \emph{true}-atoms representing
the true state of the world and a collection of \emph{knows}-atoms representing the
knowledge of the agents. It is mentioned in \citep{Thielscher17} that this definition is well-defined
for acyclic game descriptions. We observe that \cite{EngesserMNT18} proved that GDL-III is equivalent to an extension 
of DEL which allows for a succinct representation of events in update models and conditional effects.  

Comparing to GDL-III,  \mastar{} differs in the following ways. First, a GDL-III game initial state
contains only fluents about the state of the world, i.e., it only consider initial state
representing by a set of statements of the form \eqref{init1} in which $\varphi$ is a fluent. 
\mastar{} does allow arbitrary formulae. Finitary {\bf S5}-theories would subsume the set 
of initial states permissible in GDL-III.  Second, nested knowledge could be
specified in GDL-III but at the cost of extra variables. Furthermore, observation tokens
of GDL-III are about the real state of the world and thus it will also require some
extra variables (e.g, defined predicate for \emph{kwhether}) to  specify
something likes \emph{B knows that A looks into the box and knows which side of the coin is up}.
The third and perhaps most significant difference between GDL-III and \mastar{} lies in that in GDL-III, 
agents receive their individual perceptions after each round and these perceptions are true information 
about the real state of the world while agents' perceptions are dictated by the observability statements 
in \mastar{}. As such, an agent might not know the truth value of a proposition but never 
has false beliefs about the world in GDL-III while it is possible in \mastar{}. We believe that 
this stems from the fact that GDL-III focuses on knowledge of agents and \mastar{} deals with beliefs 
of agents.

\subsection{\mastar{} and Action Languages for Single-Agent Domains}

  \mastar{} is a high-level action language
for multi-agent domains. It is therefore instructive to discuss the connection
between \mastar{} and action languages for single-agent domains. First, let us observe
that \mastar{} has the following multi-agent domain specific features:
\begin{itemize}
\item it includes announcement actions; and
\item it includes specification of the agents' observability of action occurrences.
\end{itemize}
As it turns out, if we remove all features that are specific to multi-agent domains from \mastar{},
and consider the {\bf S5}-entailment as its semantics, then
the language is equivalent to the language $\cala_K$ from \cite{SonB01}.
Formally, let us consider a \mastar{} definite action theory $(I,D)$ over the signature $\langle \agents, \fluents, \cala\rangle$
such that $|\agents|=1$ and $D$ does not contain statements of the form \eqref{announce}
(announcement actions)
and statements of the form \eqref{observes}-\eqref{pobserves}. Let us  define
$$I_{\cala_K} = \{\varphi \mid \varphi \textnormal{ appears in a statement of the form } \eqref{init1}
\textnormal{ or } \eqref{init21}
\textnormal{ in } I\}.$$ Then, the following holds
\[
(I,D) \models_{\mathbf{S5}} \varphi \after \delta \:\:\:\textnormal{ iff } \:\:\:
 (I_{\cala_K},D) \models_{\cala_K} \varphi \after \delta.
\]
This shows that \mastar{} is indeed
a  generalization of action languages for single-agent domains to multi-agent domains.
This also supports the claim that other elements that have been considered in action languages of
single-agent domains, such as static causal laws, non-deterministic actions, or parallel actions could potentially
be generalized to
\mastar{}. 

\section{Conclusions and Future Works}
\label{conclusion}

In this paper, we developed an action language for representing and
reasoning about effects of actions in multi-agent domains.
The language considers world-altering actions, sensing actions, and announcement
actions. It also allows  the dynamic specification of agents'
observability with respect to
action occurrences, enabling varying degrees of visibility
of action occurrences and action effects.
The semantics of the language
relies on the notion of states (pointed Kripke structures), used
as representations of the states of the world and states
of agents' knowledge and beliefs; the semantics builds
on a transition function, which maps pairs of states
and actions to sets of states and employs the well-known notion of 
update models as the underlying machineries.

We discussed several properties of the transition function and
identified a class of theories (\emph{definite action theories})
whose set of initial {\bf S5}-states is finite, thus allowing
for the development of algorithms for the {\bf S5}-entailment relation
that is critical in applications such as planning, temporal
reasoning, and diagnosis. 

The development of \mastar{} is a first step towards the goal of
developing  scalable and efficient automated reasoning and planning systems in
multi-agent domains. Important next steps include extending the language to deal with
lying and/or misleading actions, refining the distinction
between  knowledge and beliefs of the agents, and specifying
more general models of agents' observability, to capture some of
the capabilities of update models that are missing from \mastar{}.

\section*{Acknowledgments}
The last two authors have been partially supported by NSF grants HRD-1914635, OIA-1757207, IIS-1812628.  

\bibliographystyle{elsarticle-harv}

\bibliography{../../../bibtex/bibfile,../../../bibtex/bib2010,../../../bibtex/enrico}

\section*{Appendix A: Proofs of Theorems}

\setcounter{theorem}{1}

Recall that the following notations are used in the presentation of the theorems.  
\begin{list}{$\bullet$}{\topsep=1pt \parsep=0pt \itemsep=1pt} 
\item $D$ denotes a consistent \mastar{} domain;
\item $(M,s)$ denotes a state;  and 
\item ${\sf a}$ is an action instance, 
whose precondition is given by the statement $$\executable {\sf a} \iif \psi$$ in $D$, 
and ${\sf a}$  is executable in $(M,s)$. 
\item $\rho = (F,P,O)$ is the frame of reference of the execution of ${\sf a}$ in $(M,s)$ where  
$F = F_D({\sf a}, M,s)$, 
$P = P_D({\sf a}, M,s)$, 
and  $O = O_D({\sf a}, M,s)$.
\end{list}

\begin{theorem}
Assume that ${\sf a}$ is an ontic-action instance. 
It holds that:
\begin{enumerate}
    \item for every agent $x \in F_D({\sf a}, M,s)$ and 
    $[{\sf a} \causes \ell \iif \varphi]$ belongs to $D$,  
    if $(M,s) \models \B_x \varphi$ then $\Phi_D({\sf a}, (M,s)) \models \B_x \ell$; 

    \item for every agent $y \in O_D({\sf a}, M,s)$ and a belief formula $\eta$, 
    $\Phi_D({\sf a}, (M,s)) \models \B_y \eta$ iff $(M,s) \models \B_y \eta$; 
   
   \item for every pair of agents $x \in F_D({\sf a}, M,s)$ and $y \in O_D({\sf a}, M,s)$ and a belief formula $\eta$, 
    if $(M,s) \models \B_x \B_y \eta$ then  $\Phi_D({\sf a}, (M,s)) \models \B_x \B_y \eta$. 
 
\end{enumerate}
\end{theorem}
\begin{proof}
Since ${\sf a}$ is executable in $(M,s)$, we have that $(M,s) \models \psi$. 
This means that 
\[
\Phi_D({\sf a}, (M,s)) =  (M,s) {\otimes} (\omega({\sf a},  \rho), \{\sigma\}) 
\]
where $(\omega({\sf a},  \rho), \{\sigma\})$ is given in Definition~\ref{upd-wa}.
Assume that  $(M',s') \in \Phi_D({\sf a}, (M,s))$. By Definition~\ref{def:updo}, 
we have $s' = (s,\sigma)$. 
Assume that the fluent in $\ell$ is $p$, i.e., $\ell = p$ or $\ell = \neg p$.

\begin{enumerate}

\item  
Let $\Psi^+(p,{\sf a}) = \bigvee \{ \varphi\:|\: [ {\sf a} \causes p \iif \varphi ] \in D \}$
and 
$\Psi^-(p,{\sf a}) = \bigvee \{\varphi \:|\: [ {\sf a} \causes \neg p \iif \varphi ] \in D \}$
and
$\theta = \Psi^+(p,{\sf a}) \vee (p \wedge \neg \Psi^-(p,{\sf a}))$.   
By Definition~\ref{upd-wa}, $p \rightarrow \theta \in sub(\sigma)$. 
Furthermore,  for every $u' \in M'[S]$ such that $(s',u') \in M'[x]$, it holds that $u' = (u,\sigma)$ for some $u \in M[S]$, $(M,u) \models \psi$, and $(s,u) \in M[x]$. Because $(M,s) \models \B_x \varphi$, we have that $(M,u) \models \varphi$. 
Consider two cases: 

\begin{itemize}

\item $\ell = p$. Then, $(M,u) \models \Psi^+(p,{\sf a})$, and hence, $(M,u) \models \theta$. So, $M'[\pi]((u,\sigma))\models p$.
\item $\ell = \neg p$.  Then, because $(M,u) \models \varphi$, the consistency of $D$ implies that  
$(M,u) \not\models \theta$. Therefore, $M'[\pi]((u,\sigma))\not\models p$, i.e., $M'[\pi]((u,\sigma)) \models \neg p$.
\end{itemize} 
Both cases imply that  $M'[\pi]((u,\sigma)) \models \ell$. This holds for every $u' \in M'[S]$ such that 
$(s',u') \in M'[x]$, which implies $(M',s') \models \B_x \ell$.   

\item By the construction of $M'$, we have the following observations:

\begin{itemize}
\item for every $u \in M[S]$ iff $(u,\epsilon) \in M'[S]$;  
\item for every $z \in \calag$, $(u,v) \in M[z]$ iff $((u,\epsilon), (v,\epsilon)) \in M'[z]$; and
\item for every $u \in M[S]$ and $p \in \calf$, 
$M'[\pi]((u,\epsilon)) \models p$ iff $(M',(u,\epsilon)) \models p$ 
because $sub(\epsilon) = \emptyset$.
\end{itemize} 
These observations allow us to conclude that for every formula $\eta$, $(M,u) \models \eta$ iff $(M',(u,\epsilon)) \models \eta$. 

Now, let us get back to the second item of the theorem.  
Consider 
$u' \in M'[S]$ such that $(s',u') \in M'[y]$. This holds 
iff there exists  $u \in M[S]$,  $(s,u) \in M[y]$, and $u' = (u,\epsilon)$. 

Since $(M,u) \models \eta$ iff $(M',(u,\epsilon)) \models \eta$ and 
this holds for every $u' \in M'[S]$ such that 
$(s',u') \in M'[y]$, we have that  $(M,s) \models \B_y \eta$  iff $(M',s') \models \B_y \eta$. 
 
%


\item Consider 
$u', v' \in M'[S]$ such that $(s',u') \in M'[x]$ and $(u',v') \in M'[y]$. 
This holds if there exist  $u,v \in M[S]$,  $(s,u) \in M[x]$ and $(u,v) \in M[y]$ 
such that $u' = (u,\sigma)$  and $v' = (v, \epsilon)$. 

Assume that $(M,s) \models \B_x \B_y \eta$. This implies that $(M,v) \models \eta$. The second item shows 
that $(M',(v,\epsilon)) \models\eta$, i.e., which implies $(M',s') \models \B_x \B_y \eta$. Since this holds 
for every $u', v' \in M'[S]$ such that 
$(s',u') \in M'[x]$ and $(u',v') \in M'[y]$, we have  
$(M',s') \models \B_x \B_y \ell$.

 
%
%
%
%
%
%
\end{enumerate}

Since $(M',s')$ is an arbitrary element in $\Phi_D({\sf a}, (M,s))$, the theorem holds. 
\end{proof}

\begin{theorem}
Assume that ${\sf a}$ is a sensing action instance 
and  $D$ contains the statement  ${\sf a} \determines f$.
It holds that:
\begin{enumerate}
%
%
%
%

    \item if $(M,s) \models f$ then $\Phi_D({\sf a}, (M,s)) \models \mathbf{C}_{F_D({\sf a}, M,s)} f$; 
    \item  if $(M,s) \models \neg f$ then $\Phi_D({\sf a}, (M,s)) \models \mathbf{C}_{F_D({\sf a}, M,s)} \neg f$; 

    \item $\Phi_D({\sf a}, (M,s)) \models \mathbf{C}_{P_D({\sf a}, M,s)} (\mathbf{C}_{F_D({\sf a}, M,s)} f \vee 
    								\mathbf{C}_{F_D({\sf a}, M,s)} \neg f)$;   

    \item $\Phi_D({\sf a}, (M,s)) \models \mathbf{C}_{F_D({\sf a}, M,s)} (\mathbf{C}_{P_D({\sf a}, M,s)} (\mathbf{C}_{F_D({\sf a}, M,s)} f \vee 
    								\mathbf{C}_{F_D({\sf a}, M,s)} \neg f))$;   

    \item  for every agent $y \in O_D({\sf a}, M,s)$ and formula $\eta$,  $\Phi_D({\sf a}, (M,s)) \models \B_y \eta$ iff $(M,s) \models \B_y \eta$; 
   
   \item   for every pair of agents $x \in F_D({\sf a}, M,s)$ and $y \in O_D({\sf a}, M,s)$ and a  formula $\eta$ 
    if $(M,s) \models \B_x \B_y \eta$  then $\Phi_D({\sf a}, (M,s)) \models \B_x \B_y \eta$.

%
%
         
\end{enumerate}
\end{theorem}
\begin{proof}
We will prove the theorem for the case $(M,s) \models f$. 
The proof of the theorem when $(M,s) \models \neg f$ is similar and is omitted here.  
Since ${\sf a}$ is executable in $(M,s)$, we have that $(M,s) \models \psi$. 
This means that 
\[
\Phi_D({\sf a}, (M,s)) =  M[F_D({\sf a}, M,s), f]  \otimes  (\omega({\sf a},  \rho), \{\sigma,\tau\})
\]
where $(\omega({\sf a},  \rho), \{\sigma,\tau\})$ is given in Definition~\ref{upd-sa}.
%
%
Let us denote $M[F_D({\sf a}, M,s), f]$ with $M^*$. 
Observe that by the definition of $M^*$, for each $x \in F$ there exists some $u \in M^*[x]$ such that $(M^*,u) \models f$.

We need to  prove Items 1, 3, 4, 5, and 6.  

Assume that  $(M',s') \in \Phi_D({\sf a}, (M,s))$. By Definition~\ref{def:updo}, we have $s' = (s,\sigma)$. 
 
\begin{enumerate}

\item\label{firstitem} \emph{Proof of the first item of the theorem.} 

To prove $(M',s') \models \mathbf{C}_{F} f$, we need to show that 
\[
(M',s') \models \B_{i_1} \B_{i_2} \ldots \B_{i_k}  f
\]
for any sequence $i_1,\ldots,i_k$ of agents in $F$, i.e., $i_j \in F$ for $j=1,\ldots,k$. 

Let $u'_1, \ldots, u'_{k+1} \in M'[S]$ such that $(s',u'_1) \in M'[i_1]$, $(u'_j, u'_{j+1}) \in M'[i_{j+1}]$ for $j=1,\ldots,k$.

Observe that for any $x \in F$ and $u' \in M'[S]$ such that $(s',u') \in M'[x]$, it holds that $u' = (u,\sigma)$ for some $u \in M^*[S]$, $(M^*,u) \models \psi \wedge f$, and $(s,u) \in M^*[x]$. 

This observation allows us to conclude that, for $u'_1, \ldots, u'_{k+1}$,  
there exist $u_1, \ldots,u_{k+1} \in M^*[S]$ such that $(s,u_1) \in M^*[i_1]$,  
$(u_j,u_{j+1}) \in M^*[i_{j+1}]$ for $j=1,\ldots, k$, 
and for every $j = 1, \ldots, k+1$,
$u'_j = (u_j,\sigma)$ and 
$u_i \models \psi \wedge f$. It is easy to see that this leads to $(M',s') \models  \mathbf{C}_{F} f$. 

\item\label{thirditem}  \emph{Proof of the third item of the theorem when $(M,s) \models f$.}   

To prove $(M',s') \models \mathbf{C}_{P} (\mathbf{C}_F f \vee \mathbf{C}_F \neg f)$, we need to show that 
\[
(M',s') \models \B_{i_1} \B_{i_2} \ldots \B_{i_k}  (\mathbf{C}_F f \vee \mathbf{C}_F \neg f)
\]
for any sequence $i_1,\ldots,i_k$ of agents in $P$, i.e., $i_j \in P$ for $j=1,\ldots,k$. 

Let $u'_1, \ldots, u'_{k+1} \in M'[S]$ such that $(s',u'_1) \in M'[i_1]$, $(u'_j, u'_{j+1}) \in M'[i_{j+1}]$ for $j=1,\ldots,k$.

Similar to the argument in the previous item and Definitions~\ref{def:updo} and~\ref{upd-sa}
allows us to conclude that, for $u'_1, \ldots, u'_{k+1}$,  
there exist $u_1, \ldots,u_{k+1} \in M^*[S]$ such that $(s,u_1) \in M^*[i_1]$,  
$(u_j,u_{j+1}) \in M^*[i_{j+1}]$ for $j=1,\ldots, k$, 
and for every $j = 1, \ldots, k+1$, either ({\em a}) 
$u'_j = (u_j,\sigma)$ and 
$u_i \models \psi \wedge f$ or ({\em b})  $u'_j = (u_j,\tau)$ and 
$u_i \models \psi \wedge \neg f$. 
This leads to two cases: 

\begin{enumerate}

\item  $u'_{k+1} = (u_{k+1},\sigma)$ and 
$u_{k+1} \models \psi \wedge f$. Then, similar to the proof in Item~\ref{firstitem}, we can show that $(M', u'_{k+1}) \models C_F f$. 

\item  $u'_{k+1} = (u_{k+1},\sigma)$ and 
$u_{k+1} \models \psi \wedge \neg f$. 
Again, similar to the proof in  Item~\ref{firstitem}, we can show that $(M', u'_{k+1}) \models C_F \neg f$. 
\end{enumerate} 
The two cases imply that $(M',s') \models \mathbf{C}_{P} (\mathbf{C}_F f \vee \mathbf{C}_F \neg f)$.


\item  \emph{Proof of the fourth item of the theorem when $(M,s) \models f$.}  

To prove $(M',s') \models  \mathbf{C}_{F} (\mathbf{C}_{P} (\mathbf{C}_F f \vee \mathbf{C}_F \neg f))$, we need to show that 
\[
(M',s') \models \B_{i_1} \B_{i_2} \ldots \B_{i_k}  (\mathbf{C}_{P} (\mathbf{C}_F f \vee \mathbf{C}_F \neg f))
\]
for any sequence $i_1,\ldots,i_k$ of agents in $F$, i.e., $i_j \in F$ for $j=1,\ldots,k$. 
This holds because we can show that for each $u' = (u,\sigma)$ such that $u \in M^*[S]$ and $(M^*,u) \models \psi \wedge f$,
$(M',u') \models \mathbf{C}_{P} (\mathbf{C}_F f \vee \mathbf{C}_F \neg f)$. The arguments for this conclusion are similar to the 
arguments used in the proof in Item~\ref{thirditem}. 
 
\item The proof of the fifth and sixth items of this theorem is similar to the proof of the second and third item of Theorem~\ref{ontic:theorem:1}, respectively. 


%
%
%
%
%
%
\end{enumerate}

Since $(M',s')$ is an arbitrary element in $\Phi_D({\sf a}, (M,s))$, the theorem holds. 
\end{proof}

\end{document}